\documentclass{article}

\PassOptionsToPackage{numbers,sort&compress}{natbib}
\usepackage[eandd, final]{neurips_2026}

\usepackage[utf8]{inputenc}
\usepackage[T1]{fontenc}
\usepackage{hyperref}
\usepackage{url}
\usepackage{booktabs}
\usepackage{array}
\usepackage{tabularx}
\usepackage{amsfonts}
\usepackage{amsmath}
\usepackage{amssymb}
\usepackage{nicefrac}
\usepackage{microtype}
\usepackage{xcolor}
\usepackage{graphicx}
\usepackage{xspace}
\usepackage{wrapfig}
\usepackage{colortbl}
\usepackage{pifont}                              \usepackage[capitalize,noabbrev]{cleveref}
\usepackage{algorithm}
\usepackage{algpseudocode}
\hypersetup{hidelinks}
\newcolumntype{Y}{>{\centering\arraybackslash}X}

\renewcommand{\footnoterule}{\kern-3pt\hrule width 0.55\textwidth\kern 2.6pt}
\makeatletter
\newcommand{\authornotetext}{  \begingroup
  \renewcommand{\thefootnote}{}  \long\def\@makefntext##1{\noindent##1}  \footnotetext{    \textsuperscript{*}Equal contribution. \quad
    \textsuperscript{\textdaggerdbl}Project lead. \quad
    \textsuperscript{\textdagger}Corresponding authors.\\[-1pt]
    \textsuperscript{\S}Work done during internship in KlingAI Research.  }  \endgroup
}
\makeatother

\newcommand{\ours}{\textsc{LongBanana}\xspace}
\newcommand{\scorer}{\textsc{LongBananaEval}\xspace}

\definecolor{oiBlue}{rgb}{0.0, 0.447, 0.698}
\definecolor{oiOrange}{rgb}{0.902, 0.624, 0.0}
\definecolor{oiGreen}{rgb}{0.0, 0.620, 0.451}
\definecolor{oiVermilion}{rgb}{0.835, 0.369, 0.0}
\definecolor{proprietary_model_color}{HTML}{FFFAFC}
\definecolor{open_source_model_color}{HTML}{FFFFFA}
\definecolor{skills_model_color}{HTML}{F8F8FF}

\algrenewcommand\algorithmicrequire{\textbf{Require:}}
\algrenewcommand\algorithmicensure{\textbf{Ensure:}}
\algrenewcommand\algorithmiccomment[1]{\hfill$\triangleright$ #1}

\title{\ours: An Expert-Verified Benchmark for Long-Context Multi-Reference Image Synthesis}

\author{  {\bfseries
    Haoxiang Cao\textsuperscript{1,7,*,\S} \quad
    Yuxuan Zhang\textsuperscript{2,3,*} \quad
    Penghui Du\textsuperscript{7,*} \quad
    Bo Li\textsuperscript{7} \quad
    Jiajiong Cao\textsuperscript{7}
  }\\
  {\bfseries
    Yingying Fan\textsuperscript{8} \quad
    Zimeng Wu\textsuperscript{7,\S} \quad
    Yifu Luo\textsuperscript{9} \quad
    Mingzhe Zheng\textsuperscript{10} \quad
    Junwen Miao\textsuperscript{5}
  }\\
  {\bfseries
    Huaisong Zhang\textsuperscript{7,9,\S} \quad
    Yang Yao\textsuperscript{11} \quad
    Yu Wang\textsuperscript{9} \quad
    Dongfu Jiang\textsuperscript{6} \quad
    Ping Nie\textsuperscript{6}
  }\\
  {\bfseries
    Wenhu Chen\textsuperscript{6,\textdagger} \quad
    Chaoqun Wang\textsuperscript{1,\textdagger} \quad
    Changqian Yu\textsuperscript{7,\textdaggerdbl} \quad
    Kelsey R. Allen\textsuperscript{2,3,4,\textdagger}
  }\\[6pt]
  {\normalfont\normalsize
    \textsuperscript{1}South China Normal University \quad
    \textsuperscript{2}University of British Columbia
  }\\
  {\normalfont\normalsize
    \textsuperscript{3}Vector Institute \quad
    \textsuperscript{4}Canada CIFAR AI Chair \quad
    \textsuperscript{5}Carnegie Mellon University
  }\\
  {\normalfont\normalsize
    \textsuperscript{6}University of Waterloo \quad
    \textsuperscript{7}KlingAI Research \quad
    \textsuperscript{8}University of Maryland
  }\\
  {\normalfont\normalsize
    \textsuperscript{9}Tsinghua University \quad
    \textsuperscript{10}HKUST \quad
    \textsuperscript{11}HKU
  }
}

\begin{document}

\maketitle
\authornotetext
\raggedbottom

\begin{abstract}
We introduce \ours{}, to our knowledge the \textbf{first expert-designed and expert-verified benchmark} for \emph{long-context multi-reference image synthesis} in the age of \textbf{agent-driven image generation}. The task requires retrieving dispersed visual evidence, rejecting distractors, binding entities and relations, and composing one coherent final image from multiple references. \ours{} contains 90 carefully designed tasks organized by a 3-by-3 taxonomy; each task includes about 32 reference images, 20 constraints, and 50 evaluation items, yielding \textbf{4{,}670 expert-specified atomic evaluation items} in total. Its graph-grounded evaluation design instantiates each task with reference images and an entity--relation graph, then converts node, edge, and whole-graph obligations into checklist items for Content Fidelity, Relational Binding, and Global Constraint Satisfaction. Reviewed expert labels include a 15\% double-labeled subset with \textbf{89.4\% exact agreement} and \textbf{0.83 quadratic weighted kappa}. We further introduce \ours{}-Gen, an agent-harness framework for long-horizon image generation; experiments show that minimal agent harnesses improve over direct generation, while a ReAct--Plan--Exec--Reflect (RPR) workflow yields harness-dependent gains by strengthening evidence grounding and repair. Together, \ours{} provides an auditable testbed for measuring dense visual evidence use and a shared target for developing stronger agent-driven image-generation systems.
Code and data are available at the project page: \url{https://long-banana.com/}.
\end{abstract}

\section{Introduction}
\label{sec:intro}
\begin{figure}[t]
    \centering
    \includegraphics[width=\linewidth]{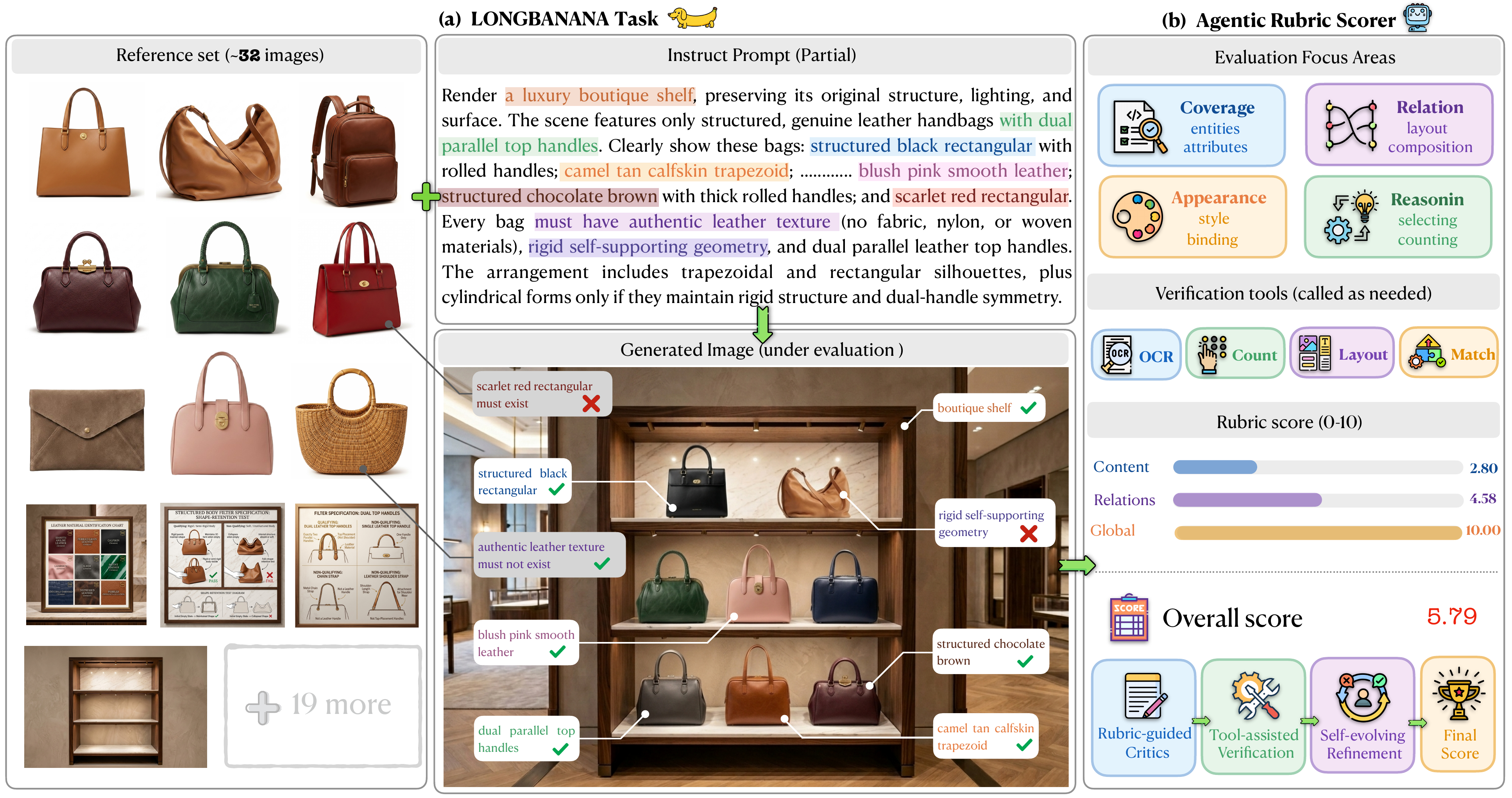}
    \caption{\ours{} task overview. (a) Each case pairs a heterogeneous reference set and structured synthesis prompt with the generated image under evaluation. (b) \ours{}-Eval converts graph-grounded checklist items into dimension-level scores and auditable evidence traces. The score labels Content, Relations, and Global denote Content Fidelity, Relational Binding, and Global Constraint Satisfaction, respectively.}
    \label{fig:task-overview}
\end{figure}

A useful image-generation system \cite{Qwen-Image,FLUX,Emu3-5,guo2025visual} should be able to synthesize a new image from the visual evidence provided by the user, even when that evidence is spread across many references. Consider an accessories retailer preparing an advertisement for all handbags with dual parallel top handles (\cref{fig:task-overview}). The system must inspect a large inventory, select the relevant bags, reject visually similar distractors, preserve appearance details, and arrange the selected items according to the requested composition. The output may look realistic while still being wrong: it may use the wrong handbag, bind a handle style to the wrong object, violate a count or layout constraint, or introduce unsupported details. The central capability is therefore not visual plausibility alone, but reference-grounded synthesis across retrieval, rejection, binding, and composition.

Evaluating this capability raises a central question: \emph{how can we tell whether an image-generation system has synthesized a correct image from long, multi-reference visual context?} The difficulty is not only that such tasks contain more images or longer prompts. The required evidence is distributed across references, entities, attributes, relations, hard negatives, unsupported-content rules, and intermediate decisions. Success can only be judged by checking whether the final image satisfies the reference-grounded obligations. A benchmark therefore needs dense multi-reference tasks and scoring that localizes errors rather than a holistic score.

Current image-generation benchmarks \cite{geditbench,emu_edit} capture important parts of this problem, but leave the full measurement target under-specified. Research on instruction-guided editing, personalization, and prompt following addresses local edits and textual requirements \cite{dpgbench,paperbanana,geditbench,imgedit,zhang2026group,cao2026dars}. Recent multi-reference benchmarks move closer to reference-based synthesis \cite{multibanana,omnicontext,multiref}, and automatic evaluation methods support systematic assessment \cite{micon,zhang2026rewardharness,zhang2026and}. What remains difficult to measure is dense evidence use: whether the model selected the right references, rejected plausible distractors, attached attributes and relations to the correct entities, and avoided content that is unsupported by either the instruction or the references. A realistic image can pass a surface-level inspection while failing exactly these obligations.

These failures are typically \emph{provenance} errors rather than rendering errors, where provenance is the source---a specific reference, or the instruction itself---to which each piece of generated content should be attributed: a system may render the right object category while copying appearance from the wrong reference, place the correct entities while violating their relation, or produce a polished composition that leaks a hard negative into the scene. To make such errors measurable, we model each task as an \emph{entity--relation graph}. Its \emph{nodes} are the entities, attributes, and text elements that must (or must not) appear; its \emph{edges} are the spatial, interaction, ordering, and local-appearance relations that bind those entities; and its \emph{whole-graph} constraints are scene-level obligations---layout, counts, temporal state, and reference selection---that no single node or edge captures. We refer to a node, an edge, or a whole-graph constraint as a \emph{graph unit}. These three levels define the three graph-grounded scoring dimensions used throughout the paper: \textit{Content Fidelity} (node level), \textit{Relational Binding} (edge level), and \textit{Global Constraint Satisfaction} (whole-graph level). We derive a \emph{checklist} of atomic, visually verifiable items, each tied to one graph unit, to localize failures.

To make this benchmark auditable at scale, we design \ours{}-Eval, a graph-grounded checklist evaluation pipeline whose scorer \scorer{} acts as an \emph{agentic checklist judge} (Section~\ref{sec:evaluator}): rather than judging the final image with one undifferentiated score, it routes each checklist item to a rubric-guided critic, invokes deterministic tools---OCR, counting, layout, and identity checks---when an item benefits from an executable check, and records evidence traces for quality control and reproducibility. We validate this measurement target on a human-audited pool of 900 generated outputs with 46{,}700 reviewed checklist labels. An independently double-labeled subset of 7{,}005 labels (15\%) reaches 89.4\% exact agreement and 0.83 quadratic weighted $\kappa$, and these reviewed expert checklist scores serve as the reference standard against which we validate \scorer{}.

We also use \ours{} to study agent-driven image-generation workflows. Each system wraps a fixed image generator in an \emph{agent harness} \cite{wu2026harnessdev,zhao2026harnessing,song2026dr}: the harness maintains reference state, selects evidence, schedules prompts, invokes the generator, and revises requests, while the image generator only synthesizes pixels. We evaluate direct generation, minimal agent harnesses, and RPR-enabled harnesses, where RPR is a serial workflow combining ReAct evidence grounding, Plan--Exec decomposition, and Reflect-based revision (Section~\ref{sec:exp}), and we compare multiple image generators under a fixed harness. Across these settings, agent harnesses improve the aggregate profile over direct generation, yet no single harness dominates: the best per-subclass scores remain distributed across direct prompting, minimal harnesses, and RPR. RPR itself yields harness-dependent gains, helping only when the harness reliably exploits retrieved evidence and revision feedback, and otherwise introducing drift.

\paragraph{Contributions}
\begin{itemize}
    \item \textbf{Benchmark setting and dataset.} We introduce \ours{}-Bench for long-context multi-reference image synthesis, with 90 expert-designed cases, dense references, hard negatives, unsupported-content rules, and a $3 \times 3$ taxonomy.
    \item \textbf{Graph-grounded evaluation.} We introduce \ours{}-Eval, mapping entity--relation graphs to atomic checklist items over Content Fidelity, Relational Binding, and Global Constraint Satisfaction.
    \item \textbf{Expert verification and audit protocol.} We provide reviewed expert checklist labels, agreement statistics, evidence traces, and scorer validation against human labels.
    \item \textbf{Agent-driven generation study.} We evaluate direct, harness-based, and RPR workflows to show that current systems fail in structured, diagnosable ways rather than along a single image-quality axis.
\end{itemize}

\section{The \ours{} Benchmark}
\label{sec:benchmark}

\subsection{Goal and Scope}
\ours{} is an evaluation benchmark for \emph{long-horizon multi-reference image synthesis}. A system is given a heterogeneous reference set and a structured synthesis instruction, and must render one coherent image that satisfies the obligations of the task's entity--relation graph (Section~\ref{sec:intro}): its required and forbidden nodes, the edges that bind them, and the whole-graph constraints on the scene. The setting differs from low-reference editing because the decisive evidence is distributed across many images, and correctness is measured by the graph-derived checklist of atomic, visually verifiable items rather than by a single holistic quality score.
\ours{} contains 90 cases: 10 cases for each subclass in a $3 \times 3$ taxonomy. A typical case contains approximately 32 references and more than 20 cross-reference constraints, and across the benchmark experts specify 4{,}670 atomic checklist items. Because every item is tied to a single graph unit, each failure is localized to a specific obligation rather than absorbed into an overall score.
\begin{figure}[t]
    \centering
    \includegraphics[width=\linewidth]{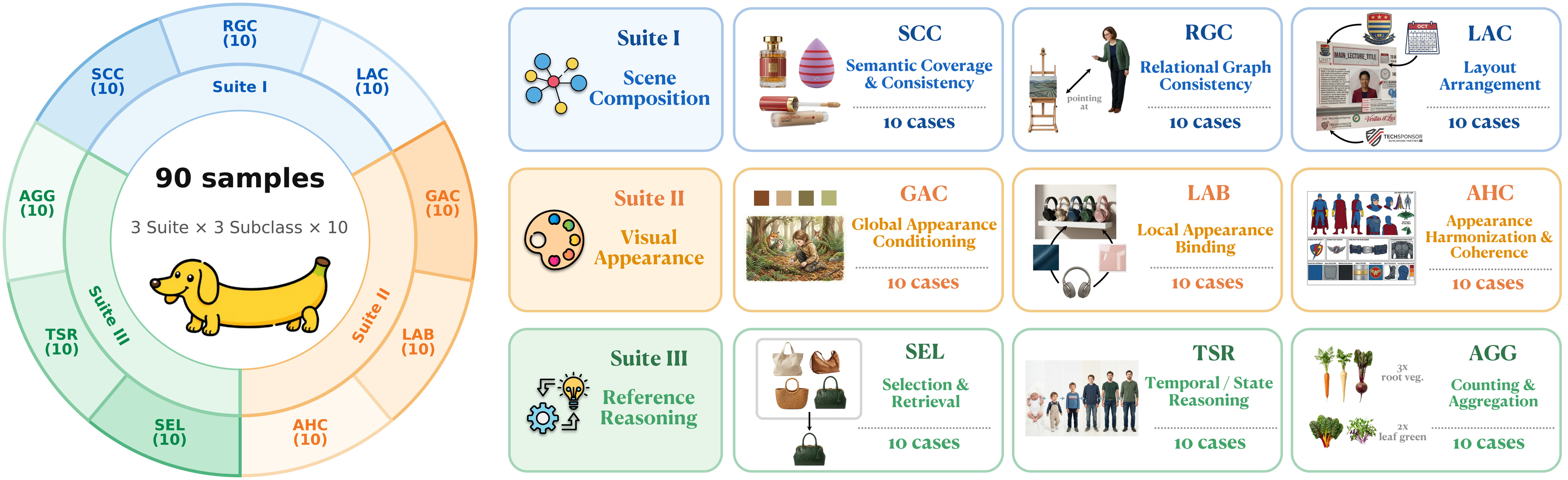}
    \caption{The $3 \times 3$ taxonomy used to organize \ours{}. Rows group cases into Scene Composition, Visual Appearance, and Reference Reasoning, while columns expose subclass-level demands used for diagnostic reporting. Each subclass contributes 10 cases.}
    \label{fig:taxonomy-grid}
\end{figure}

\subsection{Task Taxonomy and Evaluation Rubric}
\label{sec:taxonomy-rubric}
We organize \ours{} tasks into a three-suite taxonomy that captures their primary synthesis demands (\cref{fig:taxonomy-grid}). Across these tasks, a shared graph-grounded rubric evaluates whether the generated image preserves the required content, relations, and scene-level constraints.

The three suites are Scene Composition (semantic coverage, relational graph consistency, and layout arrangement), Visual Appearance (global appearance conditioning, local appearance binding, and appearance harmonization), and Reference Reasoning (selection and retrieval, temporal/state reasoning, and counting/aggregation). All tasks require a final image output; subclasses differ in the evidence-integration behavior being tested.

For each case, checklist items assess correctness at three levels of the entity--relation graph: nodes, edges, and the whole graph. \textit{Content Fidelity} measures node-level correctness: required entities, attributes, text elements, and supported semantic nodes must appear, while forbidden or unsupported nodes must be absent. \textit{Relational Binding} measures edge-level correctness, including spatial relations, interactions, ordering, support, containment, and local bindings such as texture, material, color, pose, or state assigned to a specific entity. \textit{Global Constraint Satisfaction} measures whole-graph correctness, including composition, layout, scene-level style, count and aggregation constraints, temporal/state constraints, and reference-selection constraints.

This graph-grounded rubric localizes failures to specific content, bindings, or scene-level obligations, supporting consistent diagnosis across task categories and generation systems.

\subsection{Benchmark Construction}
\begin{figure}[t]
    \centering
    \includegraphics[width=\linewidth]{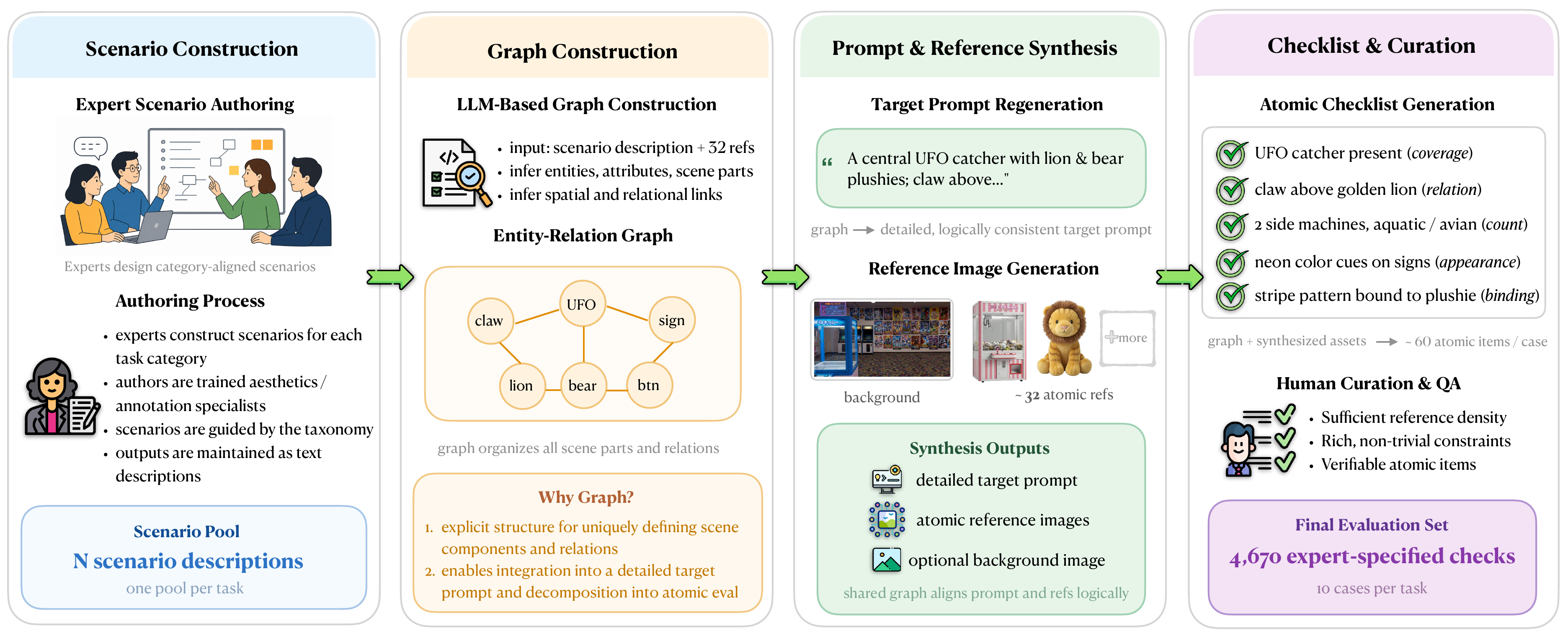}
    \caption{\ours{} construction pipeline. \textbf{(1) Scenario Construction}: experts design scenarios aligned with the task taxonomy. \textbf{(2) Graph Construction}: an LLM structures each scenario into entities, attributes, relations, and whole-scene constraints. \textbf{(3) Prompt \& Reference Synthesis}: the graph guides the creation of the target instruction and visual references. \textbf{(4) Checklist \& Curation}: graph constraints are translated into evidence-linked checklist items, and human curators review the instruction, references, and checklist for consistency, visual feasibility, and verifiability.
}
    \label{fig:benchmark-construction}
\end{figure}

We construct each case around a shared entity--relation graph that connects the synthesis instruction, reference images, and evaluation checklist (\cref{fig:benchmark-construction}). Construction proceeds from a scenario description to structured constraints, then to the visual references and instruction that express the task, and finally to the checklist used to evaluate its result.

\paragraph{Scenario and graph construction.}
Expert authors design scenarios aligned with the task taxonomy, drawing on storyboard, product, and scene-composition use cases. A scenario specifies what the final image should depict and which entities, visual attributes, and interactions must be preserved. An LLM then structures the scenario into an entity--relation graph: nodes represent entities and their attributes, edges encode spatial relations, interactions, and local appearance bindings, and whole-graph constraints specify requirements such as composition and counts. This representation makes the dependencies explicit. For example, an appearance requirement identifies both the visual property and the entity to which it belongs, while a relational requirement identifies the entities that must participate in the requested configuration. These constraints guide both task synthesis and checklist derivation.

\paragraph{Instruction and reference construction.}
The graph guides the creation of the target instruction and reference images, including entity, appearance, and optional background references. The instruction expresses the requested composition and the relationships to be realized; the references supply the visual identities, local details, and appearance cues needed to instantiate it. Discriminative visual attributes are retained in the references and omitted from text where possible. Because these cues are distributed across images, the generator must associate each relevant reference with the intended graph entity and combine their evidence while maintaining the requested relations. At least 30 cases also include two or more hard-negative references that are visually plausible but irrelevant to the requested output. Their exclusion is encoded in the checklist, allowing distractor leakage to be assessed alongside the inclusion of required content.

\paragraph{Checklist derivation.}
We translate graph constraints into atomic checklist items using the rubric introduced in Section~\ref{sec:taxonomy-rubric}. Node-level items check required entities and attributes; edge-level items check relations and appearance bindings; whole-graph items check scene-wide requirements such as layout and counts. Each item records an observable condition, the graph unit it evaluates, the relevant reference images, and its scoring criteria. This links a judgment about the output to the evidence needed to make it. Positive items test whether a required condition is satisfied, while negative items check for the absence of distractor or unsupported content. Each completed case therefore pairs an instruction and reference set with a checklist derived from the same task constraints. Section~\ref{sec:evaluator} describes how these items are scored.

\paragraph{Case curation.}
Human curators review the instruction, reference images, and checklist together, checking their correspondence to the task graph. They inspect whether the references visibly support the required identities and attributes, whether the requested relations and composition are visually feasible, and whether checklist items express clear, individually verifiable conditions. They also check reference density, constraint coverage, and overlapping or ambiguous items. Requirements that cannot be judged from the references and a candidate output are revised or rejected before inclusion. Appendix~\ref{sec:suppl-bench} provides the detailed construction protocol and case specifications. Section~\ref{sec:exp} describes human labeling of generated outputs to establish reference judgments for evaluator validation.

\section{\ours{}-Eval: Graph-Grounded Checklist Evaluation}
\label{sec:evaluator}

\begin{figure}[t]
    \centering
    \includegraphics[width=\linewidth]{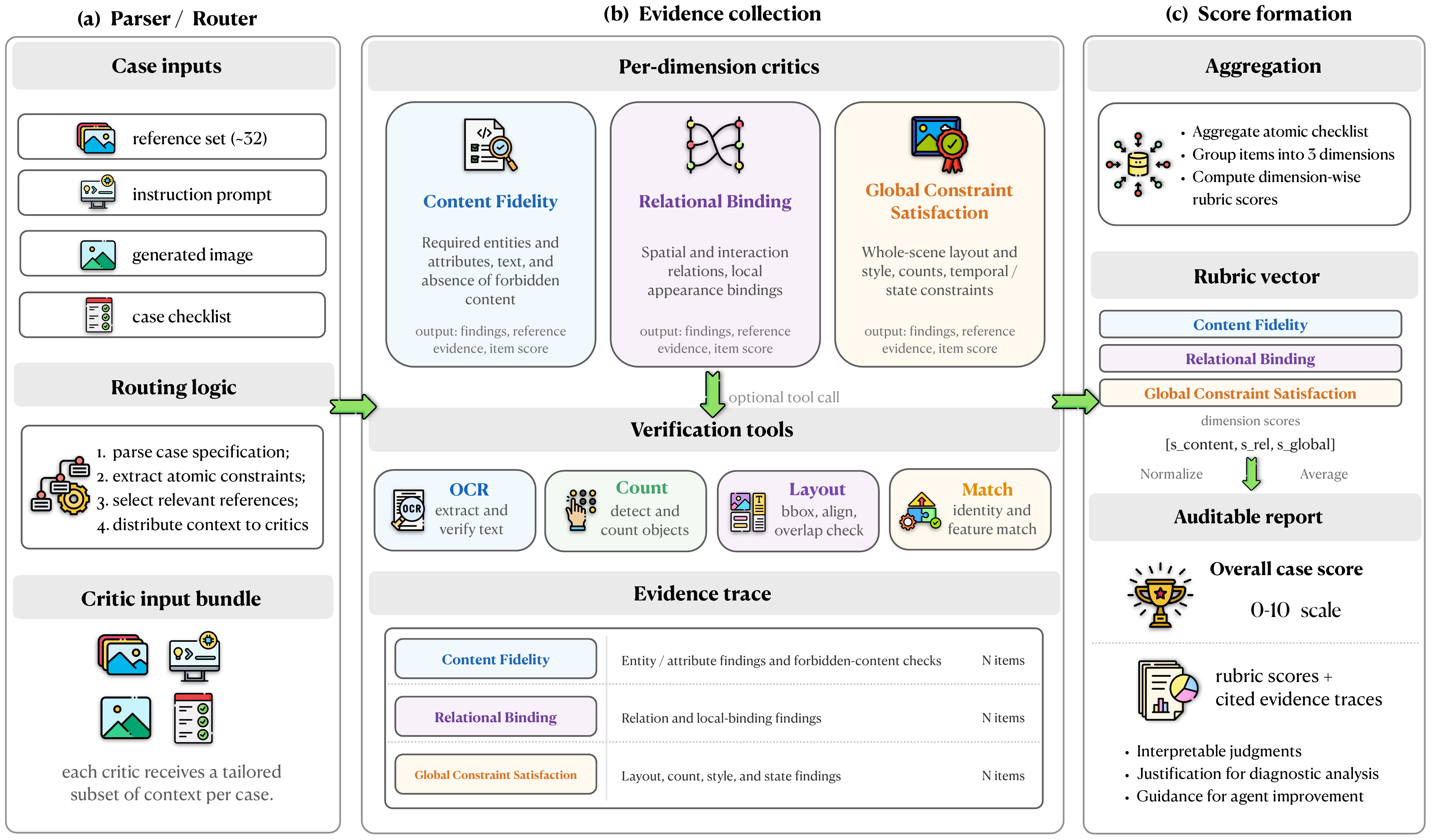}
    \caption{\ours{}-Eval scoring pipeline. The parser assembles item-specific reference evidence and routes each check to one of three rubric critics. Critics inspect the evidence, call verification tools as needed, and return item labels with supporting traces. Item labels are averaged to produce dimension-level and overall scores.}
    \label{fig:longbanana-eval}
\end{figure}

\ours{}-Eval evaluates a generated image against the case's graph-derived checklist. Its agentic checklist judge, \scorer{}, prepares item-specific reference evidence, routes each check to a dimension-specific critic, and combines the resulting judgments into dimension-level and overall scores. As shown in \cref{fig:longbanana-eval}, the pipeline also records the evidence supporting each item judgment.

\subsection{Checklist Scoring Protocol}
\paragraph{Evidence preparation and routing.}
For each item, the parser reads the condition to verify, its rubric dimension, reference identifiers, and scoring criteria from the case specification. It assembles the generated image and relevant reference images into an evidence packet, with optional crops to localize the region under inspection. Negative-content checks additionally receive the identifiers and roles of applicable hard-negative references. The router assigns the packet to the critic corresponding to the item's rubric dimension.

\paragraph{Dimension-specific checks and tools.}
The Content Fidelity critic checks required entities, attributes, and text, together with the absence of forbidden or distractor content. The Relational Binding critic checks spatial and interaction relations and the assignment of local appearance cues to their intended entities. The Global Constraint Satisfaction critic checks scene-wide requirements, including layout, global style, counts, and temporal or state constraints. Each critic compares the output with the item's reference evidence and may invoke verification tools: OCR for readable text, detection or counting for multiplicity, layout routines for spatial constraints, and identity or feature matching for consistency. The critic combines these observations with the visual evidence to determine whether the item is satisfied.

\paragraph{Item judgments and evidence traces.}
Each check produces a label $s_j\in\{0,1,2\}$: $2$ for a satisfied requirement, $1$ for partial satisfaction or uncertain evidence, and $0$ for an unsatisfied requirement. For negative items, the requirement is the absence of forbidden or unsupported content: complete absence scores $2$, ambiguous or suspicious leakage scores $1$, and clear erroneous appearance scores $0$. The evidence trace associates the label with the relevant references, the critic's findings, and any tool observations. Traces are organized by the same three rubric dimensions, linking each reported judgment to the evidence used to reach it.

\paragraph{Score aggregation.}
For a generated image with $N$ checklist items, the overall score is the mean item label scaled to $0$--$10$. Dimension scores apply the same calculation to the item set $J_d$ assigned to dimension $d$:
\[
    S_{\mathrm{overall}}=\frac{5}{N}\sum_{j=1}^{N}s_j,
    \qquad
    S_d=\frac{5}{|J_d|}\sum_{j\in J_d}s_j.
\]
The dimension scores provide a diagnostic profile, while the overall score summarizes the full checklist. The main generation tables express checklist-average scores on a $0$--$100$ scale; Section~\ref{sec:exp} separately defines the difficulty-weighted task pass rate, P@70\%.

\paragraph{Baseline and validation.}
The VLM-as-judge baseline receives the generated image, relevant references, and checklist in a single judging call and returns item labels for the same score aggregation. It shares the judging backbone and item-level criteria with \scorer{}, whose execution organizes evidence and tool use around individual checks. Section~\ref{sec:exp} evaluates both judges against reviewed human labels at the item and output-score levels; implementation details appear in Appendix~\ref{sec:suppl-eval}.

\section{Experiments}
\label{sec:exp}

Our experiments connect evaluator validation with a study of long-context image generation. We first establish agreement between automatic checklist judgments and reviewed human labels. We then ask how much explicit graph structure helps a non-agentic generator, what adaptive agentic workflows add beyond an optimized static plan, and how those gains interact with the image backend.

\subsection{Protocol and Metrics}

\paragraph{Generation conditions.}
We evaluate all systems on the held-out 90-case \ours{} benchmark. \Cref{tab:strategy_comparison} fixes Nano Banana 2 and compares two direct baselines, three minimal agent harnesses, and their RPR-enabled variants. \Cref{tab:generator_comparison_openclaw} fixes OpenClaw\,+\,RPR and compares three image backends. The shared OpenClaw\,+\,RPR with Nano Banana 2 configuration connects the two comparisons, yielding ten distinct generation conditions.

\textbf{FP-Direct} (full-prompt direct) keeps the full target prompt fixed and slides an eight-reference window over the inventory. \textbf{PD-Direct} (prompt-decomposed direct) tests how far non-agentic generation can go with explicit task structure and optimal composition scheduling. We give it privileged access to the entity--relation graph used to construct each case, including entity-to-reference correspondences. Under the same eight-reference backend budget as FP-Direct, it batches graph nodes and constructs relation-aware prompts. Its prompts expose relations incident to each batch, and each generated intermediate becomes the input for the next composition step. Integer programming solves the staging schedule exactly: first maximize relations whose anchors are composed no later than their dependents, then minimize total relational stretch $\sum_{(u,v)\in E}\lvert b(u)-b(v)\rvert$, where $b(\cdot)$ is the batch index. PD-Direct thus exploits the construction graph through a fixed, optimized execution plan; the agentic systems do not receive this graph or the evaluation checklist.

The agentic conditions use Hermes-Agent, OpenCode, and OpenClaw, with Claude Sonnet 4.6 as their shared agent model. A minimal harness maintains reference state, organizes prompts, calls the image backend, and records revisions. RPR coordinates three stages: ReAct retrieves and reasons over reference evidence; Plan--Exec decomposes the grounded request into generation or editing steps; and Reflect lets the agent inspect drafts and decide on further retrieval or revision. Appendix~\ref{sec:suppl-exp} provides the direct-baseline and workflow implementations.

\paragraph{Human reference labels.}
The human evaluation pool (the \emph{audit pool}) covers all 90 benchmark cases under 10 generation conditions, yielding $90\times10=900$ generated outputs. The case-specific checklists contain 4{,}670 items in total across the 90 cases. Applying each case's checklist to its outputs under all 10 conditions produces $4{,}670\times10=46{,}700$ item-level judgments, for which human annotators provide reviewed labels.

Human labels undergo primary annotation, second-pass review, and expert adjudication. To assess annotator consistency, 15\% of the item-level judgments---$46{,}700\times0.15=7{,}005$---receive independent labels from two annotators. On this subset, exact label agreement is 89.4\%, and quadratic weighted $\kappa$ is 0.83. Appendix~\ref{sec:suppl-bench} provides the annotation protocol, audit-correction results, and rank-stability analyses.

\paragraph{Generation scoring.}
The subclass and Overall Avg. columns measure the extent of constraint satisfaction through the unweighted checklist-average score. Each item label $s_j\in\{0,1,2\}$ is normalized as $r_j=s_j/2$, and the mean normalized label over the relevant checklist items is expressed on a $0$--$100$ scale. The subclass columns follow the nine task categories; Content Fidelity, Relational Binding, and Global Constraint Satisfaction provide the complementary item-level grading axis.

P@70\% measures task-level coverage: the percentage of the 90 case--subclass tasks whose difficulty-weighted score reaches 70. Each case contributes one task in its primary subclass. For strategy $a$ and task $t$,
\[
    \mathrm{TaskScore}(a,t)
    = 100\cdot\frac{\sum_{j\in t}w_jr_{a,j}}{\sum_{j\in t}w_j}.
\]
For the main comparisons, each item's difficulty weight $w_j$ is computed once from human scores for a fixed reference-system pool, versioned, and held fixed for subsequent systems. The overall average therefore summarizes checklist fulfillment, while P@70\% highlights how often a system reaches the task-level threshold with greater emphasis on difficult requirements. Together, they reveal whether an improvement is spread across checklist items or carries more tasks past the threshold. Appendix~\ref{sec:suppl-generation-scoring} specifies the weighting rule and aggregation details.

\subsection{Evaluator Validation}

\paragraph{Judge--human agreement.}
\Cref{tab:agreement_main} compares the single-pass VLM-as-judge baseline and \scorer{}, our agentic checklist judge, against the 900 audit-pool outputs with 46{,}700 reviewed human labels. Both judges use Gemini 3 Flash and produce item labels under the scoring protocol of Section~\ref{sec:evaluator}. We report Pearson, Spearman, and Kendall correlations between judge checklist-derived overall scores $\hat{s}_o$ and reviewed expert overall scores $s_o$.
Because this pool covers every case and every reported generation condition, these agreement measures test whether automatic judges preserve expert score trends across the complete reported output set, and provide the validation basis for reporting \scorer{} machine scores in \cref{tab:strategy_comparison,tab:generator_comparison_openclaw}.

\begin{table}[htbp]
  \centering
  \caption{Judge--human agreement on the fully covered 90-case \ours{} audit pool. Correlations compare each judge's checklist-average output-level Overall score with the reviewed human Overall score; higher values indicate closer alignment.}
  \label{tab:agreement_main}
  \small
  \setlength{\tabcolsep}{8pt}
  \begin{tabular}{lccc}
    \toprule
    Judge & Pearson & Spearman & Kendall \\
    \midrule
    VLM-as-judge & 0.8413 & 0.8248 & 0.6437 \\
    \rowcolor{oiBlue!10}
    \scorer{} (agentic checklist judge) & 0.8707 & 0.8602 & 0.6727 \\
    \bottomrule
  \end{tabular}
\end{table}

The agentic checklist judge achieves higher agreement on all three output-level correlation measures. At the item level, it also increases exact ternary agreement by 15.82 percentage points and reduces opposite-end $0\leftrightarrow2$ errors by 19.28 points (\cref{tab:checklist_executor_control}). Output-level correlations establish closer alignment with expert score trends, while the item-level comparison shows improved decisions about individual requirements. These complementary results support using \scorer{} to measure generation performance and produce evidence-linked diagnostic reports.

\subsection{Generation Results and Analysis}

\newcommand{\modellogo}[1]{\raisebox{-0.15ex}{\includegraphics[height=1em]{figures/logos/#1}}~}

\begin{table}[t]
  \centering
  \caption{Harness and workflow comparison on the held-out 90-case \ours{} benchmark. The foundation image model is fixed to Nano Banana 2, and the compared component is the generation harness/workflow. Subclass columns and Overall Avg. report the unweighted checklist-average score (0--100); P@70\% is the difficulty-weighted task pass rate. \textbf{Higher is better for all reported metrics.} Boldface and underlining mark the best and second-best values in each column.}
  \label{tab:strategy_comparison}
  \scriptsize
  \setlength{\tabcolsep}{2.0pt}
  \begin{tabularx}{\linewidth}{@{}>{\raggedright\arraybackslash}p{0.22\linewidth}*{11}{Y}@{}}
    \toprule
    \shortstack[l]{Image generator=\\Nano Banana2} & \multicolumn{3}{c}{Scene Composition}
    & \multicolumn{3}{c}{Visual Appearance}
    & \multicolumn{3}{c}{Reference Reasoning}
    & \multicolumn{2}{c}{Overall} \\
    \cmidrule(lr){2-4}
    \cmidrule(lr){5-7}
    \cmidrule(lr){8-10}
    \cmidrule(lr){11-12}
    Strategy $\downarrow$
    & SCC & RGC & LAC
    & GAC & LAB & AHC
    & SEL & TSR & AGG
    & Avg. & P@70\% \\
    \midrule
    \rowcolor{proprietary_model_color}
    \multicolumn{12}{c}{\itshape Direct generation with Nano Banana 2} \\
    \midrule
    FP-Direct & 69.6 & 61.8 & 62.0 & 88.7 & 77.5 & 71.1 & 56.4 & 73.4 & 55.1 & 68.4 & 24.4 \\
    PD-Direct & 69.9 & 61.9 & \underline{62.6} & \textbf{98.8} & 87.0 & 72.1 & 68.5 & 67.2 & 83.6 & 74.6 & 35.6 \\
    \midrule
    \rowcolor{open_source_model_color}
    \multicolumn{12}{c}{\itshape Agent harness + Nano Banana 2} \\
    \midrule
    \modellogo{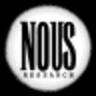}Hermes-Agent & 70.1 & 61.5 & 57.9 & 93.2 & 87.2 & 65.4 & 72.1 & 75.1 & 83.1 & 73.9 & \underline{43.3} \\
    \modellogo{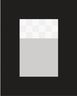}OpenCode & 71.9 & 62.3 & 62.1 & 96.3 & \textbf{90.1} & 74.2 & 75.4 & \underline{77.8} & 85.2 & 77.3 & \textbf{47.8} \\
    \modellogo{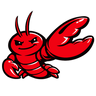}OpenClaw & 71.8 & 63.3 & 60.8 & 96.8 & 88.3 & 74.7 & 70.3 & \textbf{78.7} & 80.3 & 76.1 & 40.0 \\
    \midrule
    \rowcolor{skills_model_color}
    \multicolumn{12}{c}{\itshape Agent harness + RPR + Nano Banana 2} \\
    \midrule
    \modellogo{nous-logo.png}Hermes-Agent + RPR & 72.0 & 64.0 & 61.0 & 94.5 & 85.5 & 69.5 & 71.9 & 76.1 & \underline{86.2} & 75.6 & 35.6 \\
    \modellogo{opencode-logo.png}OpenCode + RPR & \textbf{76.3} & \textbf{64.6} & \textbf{64.1} & 92.6 & 86.8 & \textbf{78.2} & \underline{75.5} & 76.5 & 85.3 & \underline{77.8} & \underline{43.3} \\
    \rowcolor{oiBlue!10}[2\tabcolsep][2\tabcolsep]
    \modellogo{openclaw-logo.png}OpenClaw + RPR & \underline{76.1} & \underline{64.5} & \textbf{64.1} & \underline{97.1} & \underline{88.6} & \underline{75.6} & \textbf{79.8} & 73.6 & \textbf{86.4} & \textbf{78.4} & \textbf{47.8} \\
    \bottomrule
  \end{tabularx}
\end{table}

\paragraph{Generation result pattern.}
\Cref{tab:strategy_comparison} establishes PD-Direct as a strong non-agentic baseline. Graph-informed staging substantially improves on FP-Direct, with the best Global Appearance Conditioning (GAC) score and the second-best Layout Arrangement Consistency (LAC) score across all compared systems. Its improvements also extend to local appearance binding, reference selection, and aggregation. These results demonstrate the practical value of exploiting known task structure: organizing entities and their relations into a precomputed composition schedule can support competitive multi-reference synthesis. The direct baseline remains competitive on these subclasses even against systems equipped with retrieval and reflection. PD-Direct's strengths make the subsequent agentic comparison meaningful, since its task structure is supplied directly and its scheduling objective is solved exactly.

Agentic systems nevertheless achieve broader success without direct access to the construction graph. Most agentic configurations exceed PD-Direct in overall average, and OpenClaw\,+\,RPR surpasses it on eight of nine subclasses, with a 3.8-point average-score gain and a 12.2-percentage-point gain in task pass rate. This advantage reaches beyond scene composition to reference selection, state reasoning, and aggregation. The comparison exposes a gap between knowing the intended constraint structure and realizing it through successive image-generation steps. The graph specifies desired relations, while intermediate images reveal which constraints have actually survived the latest generation step. A fixed graph schedule organizes what should be composed; adaptive workflows can additionally inspect evidence, maintain reference bindings, and revise execution as synthesis unfolds. Their advantage over a structurally privileged baseline provides strong evidence for the value of agentic orchestration in long-context image generation.

Within the agentic configurations, RPR gives OpenClaw the strongest overall average and a joint-best task pass rate. Its clearest improvements over the minimal harness occur in scene composition, reference selection, and aggregation, consistent with a workflow that links evidence acquisition to execution and repair. The minimal OpenClaw harness remains stronger on temporal and state reasoning. OpenCode and Hermes-Agent also improve their average scores with RPR, while their task pass rates decline. These profiles identify reliable coordination as the next design priority: retrieved evidence must remain attached to the correct entities and constraints through planning, generation, and revision.

\paragraph{Workflow component analysis.}
The RPR ablation in \cref{tab:openclaw_component_ablation} isolates retrieval, planning, and reflection with OpenClaw and Nano Banana 2 fixed. Plan--Exec alone is strongest on global appearance conditioning and local appearance binding; ReAct alone improves reference selection over the minimal harness; and Reflect alone improves layout, aggregation, and task pass rate. The full workflow achieves the strongest overall result and pass rate among the component variants.

These profiles give a concrete direction for agent design. Evidence retrieval supplies the relevant references, planning organizes their use, and reflection checks the realized composition against the task. Their combined workflow turns these operations into a coordinated generation process, extending the benefits of decomposition with evidence inspection and feedback. The component results reinforce the main comparison: static organization is useful, and its integration with grounding and repair produces the strongest overall configuration in this study.

\begin{table}[t]
  \centering
  \caption{Foundation image-model comparison on the same 90-case \ours{} benchmark. The generation harness/workflow is fixed to OpenClaw + RPR, and the compared component is the foundation image model. Subclass columns and Overall Avg. use the same unweighted checklist-average score as \cref{tab:strategy_comparison} (0--100); P@70\% is the difficulty-weighted task pass rate. \textbf{Higher is better for all reported metrics.} Boldface and underlining mark the best and second-best values in each column.}
  \label{tab:generator_comparison_openclaw}
  \scriptsize
  \setlength{\tabcolsep}{2.0pt}
  \begin{tabularx}{\linewidth}{@{}>{\raggedright\arraybackslash}p{0.22\linewidth}*{11}{Y}@{}}
    \toprule
    \shortstack[l]{Strategy =\\OpenClaw + RPR} & \multicolumn{3}{c}{Scene Composition}
    & \multicolumn{3}{c}{Visual Appearance}
    & \multicolumn{3}{c}{Reference Reasoning}
    & \multicolumn{2}{c}{Overall} \\
    \cmidrule(lr){2-4}
    \cmidrule(lr){5-7}
    \cmidrule(lr){8-10}
    \cmidrule(lr){11-12}
    Image generator $\downarrow$
    & SCC & RGC & LAC
    & GAC & LAB & AHC
    & SEL & TSR & AGG
    & Avg. & P@70\% \\
    \midrule
    \modellogo{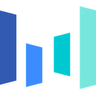}Seedream 4 & \underline{60.0} & \underline{52.6} & \underline{51.2} & 92.3 & 82.0 & 46.6 & 50.9 & \underline{67.8} & \underline{68.7} & \underline{63.6} & \underline{18.9} \\
    \modellogo{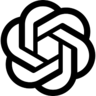}GPT Image 1 & 54.3 & 47.1 & 49.3 & \underline{94.0} & \underline{87.2} & \underline{47.4} & \underline{54.0} & 62.1 & 60.5 & 61.8 & 17.8 \\
    \rowcolor{oiBlue!10}[2\tabcolsep][2\tabcolsep]
    \modellogo{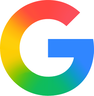}Nano Banana 2 & \textbf{76.1} & \textbf{64.5} & \textbf{64.1} & \textbf{97.1} & \textbf{88.6} & \textbf{75.6} & \textbf{79.8} & \textbf{73.6} & \textbf{86.4} & \textbf{78.4} & \textbf{47.8} \\
    \bottomrule
  \end{tabularx}
\end{table}

\paragraph{Image-backend comparison.}
With OpenClaw\,+\,RPR fixed, Nano Banana 2 leads across all nine subclasses (\cref{tab:generator_comparison_openclaw}). Seedream 4 is stronger than GPT Image 1 on scene composition, temporal and state reasoning, and aggregation, whereas GPT Image 1 is more competitive on visual appearance and reference selection. The fixed workflow makes these differences informative about each backend's ability to realize reference-grounded instructions.

Backend capability and agentic orchestration offer complementary paths to improvement. The harness controls evidence selection, constraint organization, and revision, while the image model determines how faithfully these decisions become pixels. A stronger workflow improves the use of the available visual evidence; a stronger backend improves its realization in the generated image. The broad Nano Banana 2 advantage shows that backend capability remains consequential even under the strongest evaluated workflow.

\paragraph{Diagnostic implications.}
The comparison with PD-Direct highlights the distinction between planning a composition and controlling its realization. PD-Direct uses privileged graph information to optimize reference grouping and ordering, then executes that schedule in stages. The resulting images can still depart from the intended constraints as entities are added or edited. Agentic workflows can inspect intermediate outputs, revisit reference evidence, and adjust subsequent actions. Their broader advantage motivates treating the evolving visual state, alongside the task structure, as central to generation control.

The checklist turns this control problem into inspectable requirements. PD-Direct's strong appearance conditioning alongside its weaker reference selection and state reasoning directs attention to evidence use across composition steps. Item judgments and evidence traces can distinguish an incorrect reference choice from an incorrect entity binding or an unmet scene-wide requirement. For future agent designs, these diagnostics offer a basis for choosing what to revisit and what to preserve during revision. The resulting research target is evidence-guided adaptation: connect inspection of the current image to the next retrieval, generation, or repair action while preserving satisfied constraints.

\paragraph{Further analyses.}
Appendix~\ref{sec:suppl-exp} reports the workflow ablation and case-paired score diagnostics. The generation comparisons, component study, and item-level traces together connect aggregate performance to the decisions made during synthesis, providing a basis for developing more capable reference-grounded image-generation agents.

\section{Conclusion}
\label{sec:conclusion}
\ours{} establishes long-context multi-reference image synthesis as a benchmark for integrating dispersed visual evidence, rejecting distractors, and satisfying coupled content and relational constraints. \ours{}-Bench provides 90 expert-designed and expert-verified cases across a $3 \times 3$ taxonomy, with 4{,}670 atomic checklist items. \ours{}-Eval turns these constraints into evidence-linked judgments, validated against human labels, to measure constraint fulfillment and localize generation failures.

The generation study highlights the distinction between optimizing a composition plan and adaptively controlling its realization. PD-Direct combines privileged access to the construction graph with exactly optimized staging, establishing a competitive non-agentic baseline. Agentic workflows achieve broader success without direct access to that graph, showing the value of adaptive evidence use and generation control beyond a fixed schedule. The RPR component study further supports coordinating retrieval, execution, and repair within one workflow. These findings point toward agents that reason over both task structure and the evolving visual state, preserve reference bindings across steps, and target unmet requirements during revision.

\clearpage
\begin{ack}
This work was supported by the Guangdong Basic and Applied Basic Research Foundation (Grant No. 2024A1515140109), the Youth S\&T Talent Support Programme of the Guangdong Provincial Association for Science and Technology (Grant No. SKXRC2026779), and KlingAI Research.

The authors declare no competing interests beyond the affiliations and funding disclosed in this paper.
\end{ack}

\bibliographystyle{plainnat}
\bibliography{refs}

@article{Emu3-5,
  author       = {Yufeng Cui and
                  Honghao Chen and
                  Haoge Deng and
                  Xu Huang and
                  Xinghang Li and
                  Jirong Liu and
                  Yang Liu and
                  others},
  title        = {Emu3.5: Native Multimodal Models are World Learners},
  journal      = {CoRR},
  volume       = {abs/2510.26583},
  year         = {2025}
}

@article{FLUX,
  author       = {Black Forest Labs and
                  Stephen Batifol and
                  Andreas Blattmann and
                  Frederic Boesel and
                  Saksham Consul and
                  others},
  title        = {{FLUX.1} Kontext: Flow Matching for In-Context Image Generation and
                  Editing in Latent Space},
  journal      = {CoRR},
  volume       = {abs/2506.15742},
  year         = {2025}
}

@article{Qwen-Image,
  author       = {Chenfei Wu and
                  Jiahao Li and
                  Jingren Zhou and
                  Junyang Lin and
                  Kaiyuan Gao and
                  Kun Yan and
                  Shengming Yin and
                  others},
  title        = {Qwen-Image Technical Report},
  journal      = {CoRR},
  volume       = {abs/2508.02324},
  year         = {2025}
}

@inproceedings{agentic,
  title={Agentic reward modeling: Integrating human preferences with verifiable correctness signals for reliable reward systems},
  author={Peng, Hao and Qi, Yunjia and Wang, Xiaozhi and Yao, Zijun and Xu, Bin and Hou, Lei and Li, Juanzi},
  booktitle={Proceedings of the 63rd Annual Meeting of the Association for Computational Linguistics (Volume 1: Long Papers)},
  pages={15934--15949},
  year={2025}
}

@article{agreement_metric_selection,
  title={Counting on Consensus: Selecting the Right Inter-annotator Agreement Metric for NLP Annotation and Evaluation},
  author={James, Joseph},
  journal={arXiv preprint arXiv:2603.06865},
  year={2026}
}

@inproceedings{anyedit,
  title={Anyedit: Mastering unified high-quality image editing for any idea},
  author={Yu, Qifan and Chow, Wei and Yue, Zhongqi and Pan, Kaihang and Wu, Yang and Wan, Xiaoyang and Li, Juncheng and Tang, Siliang and Zhang, Hanwang and Zhuang, Yueting},
  booktitle={Proceedings of the Computer Vision and Pattern Recognition Conference},
  pages={26125--26135},
  year={2025}
}

@misc{cao2026dars,
      title={DARS: Dual-Level Credit Assignment RL with Structured Reasoning for Instruction-Based Image Editing},
      author={Haoxiang Cao and Jiajiong Cao and Xuanpu Zhang and Changqian Yu and Chaoqun Wang},
      year={2026},
      eprint={2608.20161},
      archivePrefix={arXiv},
      primaryClass={cs.AI},
      url={https://arxiv.org/abs/2608.20161},
}

@article{creval,
  title={CREval: An Automated Interpretable Evaluation for Creative Image Manipulation under Complex Instructions},
  author={Wang, Chonghuinan and Chen, Zihan and Wei, Yuxiang and Jiang, Tianyi and Wu, Xiaohe and Li, Fan and Zuo, Wangmeng and Yao, Hongxun},
  journal={arXiv preprint arXiv:2603.26174},
  year={2026}
}

@article{deepvideo,
  title={Deep video discovery: Agentic search with tool use for long-form video understanding},
  author={Zhang, Xiaoyi and Jia, Zhaoyang and Guo, Zongyu and Li, Jiahao and Li, Bin and Li, Houqiang and Lu, Yan},
  journal={arXiv preprint arXiv:2505.18079},
  year={2025}
}

@article{dpgbench,
  title={Ella: Equip diffusion models with llm for enhanced semantic alignment},
  author={Hu, Xiwei and Wang, Rui and Fang, Yixiao and Fu, Bin and Cheng, Pei and Yu, Gang},
  journal={arXiv preprint arXiv:2403.05135},
  year={2024}
}

@inproceedings{dreambooth,
  title={Dreambooth: Fine tuning text-to-image diffusion models for subject-driven generation},
  author={Ruiz, Nataniel and Li, Yuanzhen and Jampani, Varun and Pritch, Yael and Rubinstein, Michael and Aberman, Kfir},
  booktitle={Proceedings of the IEEE/CVF conference on computer vision and pattern recognition},
  pages={22500--22510},
  year={2023}
}

@article{dreamomni2,
  title={Dreamomni2: Multimodal instruction-based editing and generation},
  author={Xia, Bin and Peng, Bohao and Zhang, Yuechen and Huang, Junjia and Liu, Jiyang and Li, Jingyao and Tan, Haoru and Wu, Sitong and Wang, Chengyao and Wang, Yitong and others},
  journal={arXiv preprint arXiv:2510.06679},
  year={2025}
}

@article{editreward,
  title={Editreward: A human-aligned reward model for instruction-guided image editing},
  author={Wu, Keming and Jiang, Sicong and Ku, Max and Nie, Ping and Liu, Minghao and Chen, Wenhu},
  journal={arXiv preprint arXiv:2509.26346},
  year={2025}
}

@article{editscore,
  title={Editscore: Unlocking online rl for image editing via high-fidelity reward modeling},
  author={Luo, Xin and Wang, Jiahao and Wu, Chenyuan and Xiao, Shitao and Jiang, Xiyan and Lian, Defu and Zhang, Jiajun and Liu, Dong and others},
  journal={arXiv preprint arXiv:2509.23909},
  year={2025}
}

@inproceedings{emu_edit,
  title={Emu edit: Precise image editing via recognition and generation tasks},
  author={Sheynin, Shelly and Polyak, Adam and Singer, Uriel and Kirstain, Yuval and Zohar, Amit and Ashual, Oron and Parikh, Devi and Taigman, Yaniv},
  booktitle={Proceedings of the IEEE/CVF Conference on Computer Vision and Pattern Recognition},
  pages={8871--8879},
  year={2024}
}

@article{geditbench,
  title={GEditBench v2: A Human-Aligned Benchmark for General Image Editing},
  author={Jiang, Zhangqi and Sun, Zheng and Zeng, Xianfang and Yang, Yufeng and Zhang, Xuanyang and Wu, Yongliang and Cheng, Wei and Yu, Gang and Yang, Xu and Wen, Bihan},
  journal={arXiv preprint arXiv:2603.28547},
  year={2026}
}

@inproceedings{gpt4_as_evaluator,
  title={Gpt-4 as an effective zero-shot evaluator for scientific figure captions},
  author={Hsu, Ting-Yao and Huang, Chieh-Yang and Rossi, Ryan and Kim, Sungchul and Giles, C and Huang, Ting-Hao},
  booktitle={Findings of the Association for Computational Linguistics: EMNLP 2023},
  pages={5464--5474},
  year={2023}
}

@inproceedings{gptscore,
  title={Gptscore: Evaluate as you desire},
  author={Fu, Jinlan and Ng, See Kiong and Jiang, Zhengbao and Liu, Pengfei},
  booktitle={Proceedings of the 2024 Conference of the North American Chapter of the Association for Computational Linguistics: Human Language Technologies (Volume 1: Long Papers)},
  pages={6556--6576},
  year={2024}
}

@misc{guo2025visual,
      title={Visual Generation Tuning},
      author={Jiahao Guo and Sinan Du and Jingfeng Yao and Wenyu Liu and Bo Li and Haoxiang Cao and Kun Gai and Chun Yuan and Kai Wu and Xinggang Wang},
      year={2025},
      eprint={2511.23469},
      archivePrefix={arXiv},
      primaryClass={cs.CV},
      url={https://arxiv.org/abs/2511.23469},
}

@misc{he2025videoscore2,
      title={VideoScore2: Think before You Score in Generative Video Evaluation},
      author={Xuan He and Dongfu Jiang and Ping Nie and Minghao Liu and Zhengxuan Jiang and Mingyi Su and Wentao Ma and Junru Lin and Chun Ye and Yi Lu and Keming Wu and Benjamin Schneider and Quy Duc Do and Zhuofeng Li and Yiming Jia and Yuxuan Zhang and Guo Cheng and Haozhe Wang and Wangchunshu Zhou and Qunshu Lin and Yuanxing Zhang and Ge Zhang and Wenhao Huang and Wenhu Chen},
      year={2025},
      eprint={2509.22799},
      archivePrefix={arXiv},
      primaryClass={cs.CV},
      url={https://arxiv.org/abs/2509.22799},
}

@misc{he2026vgi,
  title={VGI-Bench: Probing Visual Intelligence in Video Generation Models},
  author={He, Xuan and Wei, Cong and Cheng, Yuhao and Ma, Linrui and Zhang, Yuxuan and Li, Zuojun and Wen, Yuhao and Jiang, Jize and Liu, Zeyi and Hao, Yuren and Cai, Songcheng and Wu, Keming and Du, Penghui and Zou, Kai and Yang, Rui and Sun, Chenkai and Yang, Ke and Nie, Ping and Allen, Kelsey R and Wang, Chenglong and Galley, Michel and Gao, Jianfeng and Zhai, ChengXiang},
  year={2026},
  eprint={2608.19583},
  archivePrefix={arXiv},
  url={https://arxiv.org/abs/2608.19583}
}

@article{human_aligned_mllm_judges,
  title={Human-Aligned MLLM Judges for Fine-Grained Image Editing Evaluation: A Benchmark, Framework, and Analysis},
  author={Liu, Runzhou and Weingord, Hailey and Mittal, Sejal and Dungarwal, Prakhar and Nandula, Anusha and Ni, Bo and Basu, Samyadeep and Chen, Hongjie and Ahmed, Nesreen K and Li, Li and others},
  journal={arXiv preprint arXiv:2602.13028},
  year={2026}
}

@article{i2ebench,
  title={I2ebench: A comprehensive benchmark for instruction-based image editing},
  author={Ma, Yiwei and Ji, Jiayi and Ye, Ke and Lin, Weihuang and Wang, Zhibin and Zheng, Yonghan and Zhou, Qiang and Sun, Xiaoshuai and Ji, Rongrong},
  journal={Advances in Neural Information Processing Systems},
  volume={37},
  pages={41494--41516},
  year={2024}
}

@inproceedings{ice_bench,
  title={Ice-bench: A unified and comprehensive benchmark for image creating and editing},
  author={Pan, Yulin and He, Xiangteng and Mao, Chaojie and Han, Zhen and Jiang, Zeyinzi and Zhang, Jingfeng and Liu, Yu},
  booktitle={Proceedings of the IEEE/CVF International Conference on Computer Vision},
  pages={16586--16596},
  year={2025}
}

@article{ifeval,
  title={Instruction-Following Evaluation for Large Language Models},
  author={Zhou, Jeffrey and Lu, Tianjian and Mishra, Swaroop and Brahma, Siddhartha and Basu, Sujoy and Luan, Yi and Zhou, Denny and Hou, Le},
  journal={arXiv preprint arXiv:2311.07911},
  year={2023}
}

@article{imgedit,
  title={Imgedit: A unified image editing dataset and benchmark},
  author={Ye, Yang and He, Xianyi and Li, Zongjian and Lin, Bin and Yuan, Shenghai and Yan, Zhiyuan and Hou, Bohan and Yuan, Li},
  journal={arXiv preprint arXiv:2505.20275},
  year={2025}
}

@inproceedings{liu2026svcbench,
title={SVCBench: A Streaming Video Counting Benchmark for Spatial-Temporal State Maintenance},
author={Liu, Pengyiang and Shi, Zhongyue and Hao, Hongye and Fu, Qi and Bi, Xueting and Zhang, Siwei and Hu, Xiaoyang and Wang, Zitian and Huang, Linjiang and Liu, Si},
booktitle={European Conference on Computer Vision}, pages={522--538}, year={2026}, organization={Springer}
}

@inproceedings{liu2026trace,
title={TRACE: Temporal retrieval with anchored and convergent evidence for long-horizon video understanding},
author={Liu, Pengyiang and Niu, Junbo and Hu, Xiaoyang and Shi, Zhongyue and Wang, Zitian and Huang, Linjiang and Liu, Si},
booktitle={Proceedings of the 2026 Conference on Empirical Methods in Natural Language Processing},
pages={59693--59708}, year={2026}
}

@inproceedings{longbench,
  title={Longbench: A bilingual, multitask benchmark for long context understanding},
  author={Bai, Yushi and Lv, Xin and Zhang, Jiajie and Lyu, Hongchang and Tang, Jiankai and Huang, Zhidian and Du, Zhengxiao and Liu, Xiao and Zeng, Aohan and Hou, Lei and others},
  booktitle={Proceedings of the 62nd annual meeting of the association for computational linguistics (volume 1: Long papers)},
  pages={3119--3137},
  year={2024}
}

@InProceedings{lpips,
  author = {Zhang, Richard and Isola, Phillip and Efros, Alexei A. and Shechtman, Eli and Wang, Oliver},
  title = {The Unreasonable Effectiveness of Deep Features as a Perceptual Metric},
  booktitle = {Proceedings of the IEEE Conference on Computer Vision and Pattern Recognition (CVPR)},
  month = {June},
  year = {2018}
}

@article{magicbrush,
  title={Magicbrush: A manually annotated dataset for instruction-guided image editing},
  author={Zhang, Kai and Mo, Lingbo and Chen, Wenhu and Sun, Huan and Su, Yu},
  journal={Advances in Neural Information Processing Systems},
  volume={36},
  pages={31428--31449},
  year={2023}
}

@article{micon,
  title={MICON-Bench: Benchmarking and Enhancing Multi-Image Context Image Generation in Unified Multimodal Models},
  author={Wu, Mingrui and Liu, Hang and Ji, Jiayi and Sun, Xiaoshuai and Ji, Rongrong},
  journal={arXiv preprint arXiv:2602.19497},
  year={2026}
}

@article{multibanana,
  title={MultiBanana: A Challenging Benchmark for Multi-Reference Text-to-Image Generation},
  author={Oshima, Yuta and Miyake, Daiki and Matsutani, Kohsei and Iwasawa, Yusuke and Suzuki, Masahiro and Matsuo, Yutaka and Furuta, Hiroki},
  journal={arXiv preprint arXiv:2511.22989},
  year={2025}
}

@inproceedings{multiref,
  title={MultiRef: Controllable Image Generation with Multiple Visual References},
  author={Chen, Ruoxi and Chen, Dongping and Wu, Siyuan and Wang, Sinan and Lang, Shiyun and Sushko, Peter and Jiang, Gaoyang and Wan, Yao and Krishna, Ranjay},
  booktitle={Proceedings of the 33rd ACM International Conference on Multimedia},
  pages={13325--13331},
  year={2025}
}

@article{omnicontext,
  title={Omnigen2: Exploration to advanced multimodal generation},
  author={Wu, Chenyuan and Zheng, Pengfei and Yan, Ruiran and Xiao, Shitao and Luo, Xin and Wang, Yueze and Li, Wanli and Jiang, Xiyan and Liu, Yexin and Zhou, Junjie and others},
  journal={arXiv preprint arXiv:2506.18871},
  year={2025}
}

@article{paperbanana,
  title={PaperBanana: Automating Academic Illustration for AI Scientists},
  author={Zhu, Dawei and Meng, Rui and Song, Yale and Wei, Xiyu and Li, Sujian and Pfister, Tomas and Yoon, Jinsung},
  journal={arXiv preprint arXiv:2601.23265},
  year={2026}
}

@article{peng2026xdailybench,
  title={xDailyBench: Benchmarking LLMs on Professional Consultation for Real-Life Problems},
  author={Peng, Yongchang and Gu, Qingshui and Zhu, Liya and Zhang, Ge and Wang, Duo and Wang, Haodong and Ding, Jingzhe and Yu, Tianhao and Gao, Letian and Zhong, Yongjie and others},
  journal={arXiv preprint arXiv:2609.07784},
  year={2026}
}

@inproceedings{pixelreasoner,
  title={Pixel Reasoner: Incentivizing Pixel Space Reasoning via Curiosity-Driven Reinforcement Learning},
  author={Su, Alex and Wang, Haozhe and Ren, Weiming and Lin, Fangzhen and Chen, Wenhu},
  booktitle={The Thirty-ninth Annual Conference on Neural Information Processing Systems},
  year={2025}
}

@article{plcc,
  title={VII. Mathematical contributions to the theory of evolution.---III. Regression, heredity, and panmixia},
  author={Pearson, Karl},
  journal={Philosophical Transactions of the Royal Society of London. Series A, containing papers of a mathematical or physical character},
  volume={187},
  pages={253--318},
  year={1896},
  publisher={The Royal Society London}
}

@inproceedings{prometheus2,
  title={Prometheus 2: An open source language model specialized in evaluating other language models},
  author={Kim, Seungone and Suk, Juyoung and Longpre, Shayne and Lin, Bill Yuchen and Shin, Jamin and Welleck, Sean and Neubig, Graham and Lee, Moontae and Lee, Kyungjae and Seo, Minjoon},
  booktitle={Proceedings of the 2024 Conference on Empirical Methods in Natural Language Processing},
  pages={4334--4353},
  year={2024}
}

@book{psnr,
  title={Fundamentals of digital image processing},
  author={Jain, Anil K},
  year={1989},
  publisher={Prentice-Hall, Inc.}
}

@article{song2026dr,
  title={Dr. Claw: An AI Scientist Workspace for Vibe Research},
  author={Song, Dingjie and Zhang, Hanrong and Liu, Dawei and Liu, Yixin and Li, Zongxia and Yuan, Zhengqing and Zhang, Siqi and Zou, Henry Peng and Yan, Zhiling and Zhang, Yuxuan and others},
  journal={arXiv preprint arXiv:2609.00365},
  year={2026}
}

@article{spts,
  title={Probe and Skip: Self-Predictive Token Skipping for Efficient Long-Context LLM Inference},
  author={Wu, Zimeng and Wang, Donghao and Jin, Chaozhe and Chen, Jiaxin and Wang, Yunhong},
  journal={arXiv preprint arXiv:2601.13155},
  year={2026}
}

@article{srcc,
 ISSN = {00029556},
 URL = {http://www.jstor.org/stable/1412159},
 author = {C. Spearman},
 journal = {The American Journal of Psychology},
 number = {1},
 pages = {72--101},
 publisher = {University of Illinois Press},
 title = {The Proof and Measurement of Association between Two Things},
 urldate = {2026-04-16},
 volume = {15},
 year = {1904}
}

@article{ssim,
  title={Image quality assessment: from error visibility to structural similarity},
  author={Wang, Zhou and Bovik, Alan C and Sheikh, Hamid R and Simoncelli, Eero P},
  journal={IEEE transactions on image processing},
  volume={13},
  number={4},
  pages={600--612},
  year={2004},
  publisher={IEEE}
}

@inproceedings{tifa,
  title={TIFA: Accurate and Interpretable Text-to-Image Faithfulness Evaluation with Question Answering},
  author={Hu, Yushi and Liu, Benlin and Kasai, Jungo and Wang, Yizhong and Ostendorf, Mari and Krishna, Ranjay and Smith, Noah A.},
  booktitle={Proceedings of the IEEE/CVF International Conference on Computer Vision (ICCV)},
  pages={20406--20417},
  year={2023}
}

@article{uavgen,
  title={Visual Prototype Conditioned Focal Region Generation for UAV-Based Object Detection},
  author={Li, Wenhao and Wu, Zimeng and Wu, Yu and Fu, Zehua and Chen, Jiaxin},
  journal={arXiv preprint arXiv:2604.02966},
  year={2026}
}

@article{vrthinker,
  title={Vr-thinker: Boosting video reward models through thinking-with-image reasoning},
  author={Wang, Qunzhong and Liu, Jie and Liang, Jiajun and Jiang, Yilei and Zhang, Yuanxing and Zheng, Yaozhi and Wang, Xintao and Wan, Pengfei and Yue, Xiangyu and Liu, Jiaheng},
  journal={arXiv preprint arXiv:2510.10518},
  year={2025}
}

@article{wu2026harnessdev,
  title={HarnessDev: Can LLMs Create and Evolve Their Own Agent Harness?},
  author={Wu, Yuhao and Zhang, Jingyuan and Shi, Jiajun and Lei, Xinping and Gu, Qingshui and Zhang, Yuxuan and Wang, Zexuan and He, Chen and Huang, Chen and Song, Maojia and others},
  journal={arXiv preprint arXiv:2609.01437},
  year={2026}
}

@misc{yang2025structeval,
      title={StructEval: Benchmarking LLMs' Capabilities to Generate Structural Outputs},
      author={Jialin Yang and Dongfu Jiang and Lipeng He and Sherman Siu and Yuxuan Zhang and Disen Liao and Zhuofeng Li and Huaye Zeng and Yiming Jia and Haozhe Wang and Benjamin Schneider and Chi Ruan and Wentao Ma and Zhiheng Lyu and Yifei Wang and Yi Lu and Quy Duc Do and Ziyan Jiang and Ping Nie and Wenhu Chen},
      year={2026},
      eprint={2505.20139},
      archivePrefix={arXiv},
      primaryClass={cs.SE},
      url={https://arxiv.org/abs/2505.20139},
}

@misc{yao2025rigorous,
      title={Dr. Bench: A Multidimensional Evaluation for Deep Research Agents, from Answers to Reports},
      author={Yang Yao and Yixu Wang and Yuxuan Zhang and Yi Lu and Tianle Gu and Lingyu Li and Dingyi Zhao and Keming Wu and Haozhe Wang and Ping Nie and Yan Teng and Yingchun Wang},
      year={2026},
      eprint={2510.02190},
      archivePrefix={arXiv},
      primaryClass={cs.AI},
      url={https://arxiv.org/abs/2510.02190},
}

@misc{zhang2026and,
      title={Where, What, Why, and Importance: Structured Defect Grounding for Text-to-Image Feedback},
      author={Huaisong Zhang and Hao Yu and Yuxuan Zhang and Jiahe Wang and Xinrui Chen and Haoxiang Cao and Feng Lu and Wendong Zhang and Changqian Yu and Chun Yuan},
      year={2026},
      eprint={2606.06113},
      archivePrefix={arXiv},
      primaryClass={cs.CV},
      url={https://arxiv.org/abs/2606.06113},
}

@misc{zhang2026clawbench,
      title={ClawBench: Can AI Agents Complete Everyday Online Tasks?},
      author={Yuxuan Zhang and Yubo Wang and Yipeng Zhu and Penghui Du and Junwen Miao and Xuan Lu and Zhuofeng Li and Xingwei Qu and Zhengkang Guo and Yuanzhe Shen and Dingjie Song and Han Zhou and Tuney Zheng and Xian Wu and Hao Yu and Songcheng Cai and Yi Lu and Yunzhuo Hao and Minyi Lei and Liang Chen and Kai Zou and Huifeng Yin and Wendong Xu and Dongfu Jiang and Ping Nie and Jiaheng Liu and Wenhu Chen and Kelsey R. Allen},
      year={2026},
      eprint={2604.08523},
      archivePrefix={arXiv},
      primaryClass={cs.CL},
      url={https://arxiv.org/abs/2604.08523},
}

@misc{zhang2026group,
      title={Group Relative Attention Guidance for Image Editing},
      author={Xuanpu Zhang and Xuesong Niu and Ruidong Chen and Dan Song and Jianhao Zeng and Penghui Du and Haoxiang Cao and Kai Wu and An-an Liu},
      year={2025},
      eprint={2510.24657},
      archivePrefix={arXiv},
      primaryClass={cs.CV},
      url={https://arxiv.org/abs/2510.24657},
}

@misc{zhang2026rewardharness,
      title={RewardHarness: Self-Evolving Agentic Post-Training},
      author={Yuxuan Zhang and Cong Wei and Penghui Du and Bo Li and Junwen Miao and Huaisong Zhang and Songcheng Cai and Yubo Wang and Dongfu Jiang and Yuyu Zhang and Changqian Yu and Ping Nie and Wenhu Chen and Kelsey R. Allen},
      year={2026},
      eprint={2605.08703},
      archivePrefix={arXiv},
      primaryClass={cs.AI},
      url={https://arxiv.org/abs/2605.08703},
}

@article{zhao2026harnessing,
  title={Harnessing LLMs as Agents: What Does It Cost?},
  author={Zhao, Zelin and Guo, Xinyu and Zhang, Jingyuan and Zhang, Yuxuan and Chen, Yongxin},
  journal={arXiv preprint arXiv:2610.02488},
  year={2026}
}

@article{zhu2026startupbench,
  title={StartupBench: Benchmarking General-Purpose Agents on Market-Validated End-to-End Workflows},
  author={Zhu, Liya and Ma, Xin and Liu, Tao and Wang, Haodong and Zhang, Ge and Ding, Jingzhe and Gu, Qingshui and Zhong, Yongjie and Meng, Jinxiang and Gao, Yuan and others},
  journal={arXiv preprint arXiv:2608.17800},
  year={2026}
}

@article{zhu2026workflow,
  title={Workflow-gym: Towards long-horizon evaluation of computer-use agentic tasks in real-world professional fields},
  author={Zhu, Liya and Ding, Jingzhe and Zhang, Jian and Xue, Jianbo and Liang, Shihao and Zhang, Ge and Zhu, Yi and Zeng, Duju and Gao, Xiang and Gu, Qingshui and others},
  journal={arXiv preprint arXiv:2606.11042},
  year={2026}
}
\clearpage
\appendix
\clearpage
\section*{Appendix Contents}
\label{sec:appendix-contents}
\begin{enumerate}
  \item \hyperref[sec:related]{Appendix~\ref*{sec:related}: Related work.}
  Image-synthesis benchmarks, atomic constraint evaluation, and automatic judges.
  \item \hyperref[sec:suppl-bench]{Appendix~\ref*{sec:suppl-bench}: Benchmark and dataset details.}
  Taxonomy, construction rationale, benchmark statistics, provenance, machine-readable evaluation schema, and expert-label quality control.
  \item \hyperref[sec:suppl-eval]{Appendix~\ref*{sec:suppl-eval}: \scorer{} details.}
  Model and agent declarations, judge--human agreement, tool aggregation rules, and difficulty-weighted scoring.
  \item \hyperref[sec:suppl-exp]{Appendix~\ref*{sec:suppl-exp}: Additional generation experiments and diagnostics.}
  Direct baselines, agent-orchestrator ablations, and case-paired score diagnostics.
  \item \hyperref[sec:suppl-responsible-release]{Appendix~\ref*{sec:suppl-responsible-release}: Responsible release and safeguards.}
  Limitations, impact considerations, and release safeguards.
\end{enumerate}

\vspace{1em}

\section{Related Work}
\label{sec:related}

\paragraph{Image Synthesis Benchmarks.}
Image-generation benchmarks cover instruction-guided editing, personalization, human alignment, and reasoning-oriented synthesis \cite{magicbrush,emu_edit,dreambooth,i2ebench,anyedit,ice_bench} but generally focus on creating a single image from a single prompt. Recent multi-reference benchmarks \cite{multibanana,omnicontext,dreamomni2,multiref} are closer to our setting, but they typically use fewer references or lower-density cross-image constraints. \ours{} targets final-image synthesis in which correctness depends on retrieving, rejecting, binding, and integrating evidence distributed across dozens of references.

\paragraph{Atomic Constraint Evaluation.}
Long-context tasks have long been a central challenge across language and vision domains \cite{multibanana,longbench,spts,liu2026trace,liu2026svcbench}. Instruction-following benchmarks in NLP often convert open-ended tasks into verifiable atomic requirements. IFEval \cite{ifeval} emphasizes directly checkable constraints, and TIFA \cite{tifa} converts text-to-image faithfulness into question-answer checks. \ours{} transfers this principle to long-horizon multi-reference synthesis. Each checklist item links the output to a graph node, graph edge, or whole-graph constraint, so failures can be traced to missing content, incorrect relations, hard-negative leakage, unsupported content, or global constraint violations.

\paragraph{Automatic Judges.}
Low-level metrics such as PSNR, SSIM, and LPIPS \cite{psnr,ssim,lpips} are poorly matched to reference faithfulness, motivating model-based judges and rubric evaluators \cite{gptscore,gpt4_as_evaluator,prometheus2,yang2025structeval,human_aligned_mllm_judges,creval,yao2025rigorous,agentic,zhang2026clawbench,zhang2026rewardharness,editreward,he2025videoscore2,zhu2026startupbench,peng2026xdailybench,he2026vgi,zhu2026workflow}. However, a holistic judge can conflate realism with correctness when many references are present \cite{editscore,pixelreasoner,vrthinker,deepvideo,uavgen}. \scorer{} evaluates checklist-level evidence with routed critics and tools, and we report its validity through Pearson, Spearman, and Kendall agreement with reviewed expert labels \cite{srcc,plcc,agreement_metric_selection}.

\section{Benchmark and Dataset Details}
\label{sec:suppl-bench}

\subsection{Detailed Taxonomy}
\Cref{tab:taxonomy} gives the full $3 \times 3$ taxonomy used to organize \ours{}. The suites separate scene composition, visual appearance, and reference reasoning demands, while the subclasses specify the dominant checklist requirement for each group of cases.
\begin{table}[htbp]
  \centering
  \caption{The $3 \times 3$ task taxonomy of \ours{}. All tasks require synthesizing a final image from multiple references; subclasses differ in their dominant evaluative demand rather than output modality. \ours{} contains 10 cases per subclass, for 90 cases in total.}
  \label{tab:taxonomy}
  \footnotesize
  \setlength{\tabcolsep}{3pt}
  \begin{tabular}{p{2.2cm}cp{3.4cm}p{4.7cm}c}
    \toprule
    Suite & Code & Subclass & Dominant evaluative demand & Held-out cases \\
    \midrule
    Suite I & SCC & Semantic Coverage \& Consistency & Preserve all required entities, attributes, and semantic elements from the reference set. & 10 \\
    Suite I & RGC & Relational Graph Consistency & Satisfy required interactions, spatial relations, and topology-level constraints among referenced entities. & 10 \\
    Suite I & LAC & Layout Arrangement Consistency & Render the correct global composition, scale, occlusion, and foreground/background organization. & 10 \\
    Suite II & GAC & Global Appearance Conditioning & Apply scene-level appearance cues such as style, palette, lighting, or atmosphere to the whole image. & 10 \\
    Suite II & LAB & Local Appearance Binding & Bind different appearance references to specific entities, regions, or materials without cross-talk. & 10 \\
    Suite II & AHC & Appearance Harmonization \& Coherence & Merge multiple appearance references into a visually coherent and physically plausible final image. & 10 \\
    Suite III & SEL & Selection \& Retrieval & Select the correct subset of references required by the case and reflect that selection in the output. & 10 \\
    Suite III & TSR & Temporal / State Reasoning & Infer state changes, temporal order, or action outcomes from references before synthesis. & 10 \\
    Suite III & AGG & Counting \& Aggregation & Count, summarize, or aggregate evidence from references and render the result in the final image. & 10 \\
    \midrule
    \textbf{Total} & & & & \textbf{90} \\
    \bottomrule
  \end{tabular}
\end{table}

\subsection{Construction Rationale}
\Cref{tab:construction_rationale} expands the construction choices summarized in \cref{sec:benchmark}.
\begin{table}[H]
  \centering
  \caption{Construction rationale for \ours{}. Each step has an explicit input, operation, purpose, and quality-control check so that benchmark cases are grounded in auditable reference-based synthesis requirements.}
  \label{tab:construction_rationale}
  \small
  \setlength{\tabcolsep}{4pt}
  \resizebox{\linewidth}{!}{  \begin{tabular}{p{2.5cm}p{3.4cm}p{4.5cm}p{4.8cm}}
    \toprule
    Step & Input / operation & Why it is needed & Quality control \\
    \midrule
    Scenario authoring & Expert scenario description grounded in realistic storyboard, product, or scene-composition use cases & Ensures each task is visually concrete and long-horizon rather than a random prompt & Screen for reference density, scene feasibility, and taxonomy fit \\
    Graph construction & Entity and attribute nodes, relation and binding edges, and whole-graph constraints extracted from the scenario & Provides the source of truth for prompt construction and checklist generation & Human audit removes ambiguous entities, weak relations, and unverifiable constraints \\
    Reference generation & Clean reference images for objects, backgrounds, styles, local attributes, and optional distractors & Forces systems to use visual evidence rather than text shortcuts & Discriminative attributes are visible in references and omitted from text when possible \\
    Hard-negative insertion & At least 30 cases include two or more irrelevant but visually plausible references & Tests resistance to distractors, a common failure amplifier in agentic workflows & Checklists contain forbidden-use items and annotators mark distractor leakage \\
    Checklist derivation & Atomic requirements generated from graph nodes, graph edges, whole-graph constraints, counts, and forbidden content & Converts complex synthesis into auditable ordinal decisions & Items must be reference-dependent, visually verifiable, and non-overlapping where possible \\
    Reported-evaluation isolation & Case identifiers and quality-check records kept separate from the reported benchmark & Keeps benchmark design checks separate from reported results & Human labels and scorer outputs are reported for the fixed 90-case benchmark \\
    \bottomrule
  \end{tabular}}
\end{table}

\subsection{Benchmark Statistics}
\Cref{tab:dataset_stats} summarizes the fixed benchmark scale and annotation load. These quantities describe the benchmark design itself, including the number of suites, cases, references, constraints, checklist items, and hard-negative settings, rather than model performance.
\begin{table}[H]
  \centering
  \caption{\ours{} benchmark statistics by design. Entries describe the fixed benchmark specification rather than model outcomes.}
  \label{tab:dataset_stats}
  \small
  \begin{tabular}{lc}
    \toprule
    Statistic & Value \\
    \midrule
    Major suites & 3 \\
    Subclasses & 9 \\
    Cases per subclass & 10 \\
    Total cases & 90 \\
    Typical reference count per case & approximately 32 \\
    Typical cross-reference constraints per case & 20+ \\
    Mean atomic checklist items per case & 51.9 \\
    Total atomic checklist items & 4{,}670 \\
    Hard-negative robustness cases & at least 30 \\
    Hard-negative references per robustness case & at least 2 \\
    \bottomrule
  \end{tabular}
\end{table}

\subsection{Benchmark Comparison}
\Cref{tab:bench_comparison} positions \ours{} relative to representative image-editing and image-synthesis benchmarks. The comparison emphasizes why \ours{} targets long-horizon multi-reference synthesis: dense reference sets, many cross-reference constraints, atomic checklist evaluation, and explicit human-label quality control.
\begin{table}[H]
  \centering
  \caption{Comparison of \ours{} to representative image-editing and image-synthesis benchmarks. \ours{} is positioned as a high-density, expert-verifiable benchmark for long-horizon multi-reference synthesis with explicit expert-label quality-control reporting.}
  \label{tab:bench_comparison}
  \small
  \setlength{\tabcolsep}{4pt}
  \resizebox{\columnwidth}{!}{  \begin{tabular}{lccccc}
    \toprule
    Benchmark & Date & Human scale & Label QA / agreement & Code open? & Dense multi-ref? \\
    \midrule
    TIEdit              & 2026-03 & 20 raters; 307K ratings   & SRCC/PLCC/KRCC & \ding{55} & \ding{55} \\
    CREval              & 2026-03 & 18 raters; $>$200 cases   & not central    & \checkmark & \ding{55} \\
    HumanAligned-MLLM-J.& 2026-02 & 25 raters; 100 tasks      & ICC(2,k)       & \checkmark & \ding{55} \\
    GEditBench v2       & 2026-03 & experts; 3{,}506 pairs    & accuracy       & \checkmark & \ding{55} \\
    EditHF-1M           & 2026-03 & pros; 29.1M pairs         & not reported   & \checkmark & \ding{55} \\
    Co-EditBench        & 2025-09 & 34 raters; 800 cases      & SROCC/PLCC/KLCC & \ding{55} & \ding{55} \\
    \midrule
    \rowcolor{oiBlue!10}
    \ours{}             & 2026 & expert protocol; 90 cases & second/third review + audit accuracy & public release & \checkmark \\
    \bottomrule
  \end{tabular}
  }
\end{table}

\subsection{Asset Provenance, Licenses, and Consent}
Each source image or generated asset is tracked with source, creator when known, access date, intended use, redistribution status, and privacy/safety screening status. Assets with uncertain provenance, personal-data concerns, unsafe content, or incompatible redistribution terms are excluded from unrestricted release or handled through controlled access.

\subsection{Evaluation Specification Schema}
Each case serializes the evaluation specification in a machine-readable schema:
\begingroup
\raggedright
\begin{itemize}
  \setlength{\itemsep}{1pt}
  \item \textbf{Case fields:} \texttt{case\_id}, \texttt{suite}, \texttt{subclass}, \texttt{instruction}, \texttt{reference\_inventory}, \texttt{evidence\_graph} (the entity--relation graph), \texttt{hard\_negative\_references}, \texttt{unsupported\_content\_rules}, \texttt{checklist\_items}, \texttt{rubric\_dimensions}, and \texttt{provenance\_status}.
  \item \textbf{Checklist-item fields:} \texttt{item\_id}, \texttt{dimension}, \texttt{graph\_unit}, \texttt{polarity}, \texttt{required\_references}, \texttt{observable\_condition}, \texttt{label\_set}, and \texttt{human\_evidence\_notes}.
  \item \textbf{Rubric dimensions:} \texttt{content\_fidelity}, \texttt{relational\_binding}, and \texttt{global\_constraint\_satisfaction}.
\end{itemize}
\endgroup

\subsection{Checklist Item Template}
A checklist item is valid only if it is visually verifiable, reference-grounded, and non-redundant where possible. Positive items evaluate whether a required fact is present and correctly bound. Negative items are Content Fidelity checks that evaluate whether a distractor or unsupported element is absent: completely absent is labeled 2, visually suspicious or ambiguous leakage is labeled 1, and full erroneous appearance is labeled 0. Items requiring specialist interpretation include a short reference note identifying the evidence to inspect.

\subsection{Annotation Interface and Protocol}
Annotators view the reference set, target instruction, generated output, checklist item evidence, and explicit 0/1/2 scoring instructions. The human-audited multi-system output pool covers all 90 cases and all 10 reported generation conditions, yielding 900 generated outputs. Human reviewers assign 46{,}700 reviewed checklist labels by applying all 4{,}670 benchmark checklist items to every reported condition. The three rubric dimensions are computed from the corresponding checklist labels. The protocol uses primary annotation, second-pass review, third-pass expert adjudication, an independently double-labeled subset of 7{,}005 labels, stratified audit sampling, schema validation, and score-stability checks. The annotation audit records include qualification and training criteria, review status, correction logs, adjudication logs, audit outcomes, and stability summaries without exposing annotator identities. \Cref{fig:human_annotation_platform_examples,fig:human_annotation_platform_relation} show screenshots from the checklist annotation platform used for these human reviews.

\subsection{Expert Label Quality Control}
\Cref{tab:label_quality_audit} summarizes label quality control. All 46{,}700 judgments undergo review and adjudication, with difficult items receiving further review to establish final labels. Errors found in the random audit trigger expanded checks of related labels, and the label set is frozen after all revisions. \Cref{tab:score_stability} compares scores and rankings for the same ten generation conditions across label versions using the same checklist-average aggregation.
\begin{table}[H]
  \centering
  \caption{Expert annotation quality for the 90-case \ours{} benchmark. Human reviewers label all 4{,}670 benchmark checklist items for all 10 reported generation conditions, yielding 46{,}700 reviewed output-level checklist labels. The table reports the quality-control records behind those human labels, including independent double labeling, adjudication, audit, and schema validation.}
  \label{tab:label_quality_audit}
  \small
  \setlength{\tabcolsep}{6pt}
  \resizebox{\linewidth}{!}{  \begin{tabular}{lcc}
    \toprule
    Quality check & Scope & Reported value \\
    \midrule
    Held-out test cases & 90 cases & 90 \\
    Audited generation conditions & all reported generation conditions & 10 \\
    Audited generated outputs & $90 \times 10$ outputs & $900$ \\
    Benchmark checklist items & 90 cases total & $4{,}670$ \\
    Reviewed human checklist labels & $4{,}670 \times 10$ conditions & $46{,}700$ labels \\
    Independently double-labeled subset & 15\% of reviewed labels & $7{,}005$ labels \\
    Exact annotator agreement & double-labeled subset & 89.4\% \\
    Quadratic weighted $\kappa$ & double-labeled subset & 0.83 \\
    Expert adjudication rate & double-labeled subset & 9.2\% \\
    Random audit sample & final labels & $5{,}604$ labels \\
    Audit correction rate & random audit sample & 3.34\% \\
    Schema validation errors & released label table & 0 \\
    \bottomrule
  \end{tabular}  }
\end{table}

\begin{table}[H]
  \centering
  \caption{Score stability under label correction for the ten reported generation conditions. The pre-audit to post-audit correction rate uses the full 46{,}700-label pool and includes expanded checks beyond the random audit sample. Checklist-average score changes are expressed in percentage points; Kendall $\tau$ and Top-1 compare condition rankings before and after revision.}
  \label{tab:score_stability}
  \small
  \setlength{\tabcolsep}{5pt}
  \resizebox{\linewidth}{!}{  \begin{tabular}{lccccc}
    \toprule
    Comparison & Corrected labels & Mean score delta & Max score delta & Rank Kendall $\tau$ & Top-1 changed? \\
    \midrule
    Primary $\rightarrow$ Reviewed & 4.5\% & 1.2 pp & 2.8 pp & 0.96 & No \\
    Reviewed $\rightarrow$ Adjudicated & 1.8\% & 0.5 pp & 1.4 pp & 0.99 & No \\
    Pre-audit $\rightarrow$ Post-audit & 1.3\% & 0.3 pp & 0.9 pp & 1.00 & No \\
    Primary $\rightarrow$ Final & 5.7\% & 1.5 pp & 3.1 pp & 0.95 & No \\
    \bottomrule
  \end{tabular}  }
\end{table}

\subsection{Human Annotation Platform and Audit Trail}
\begin{figure}[H]
  \centering
  \includegraphics[width=.96\linewidth]{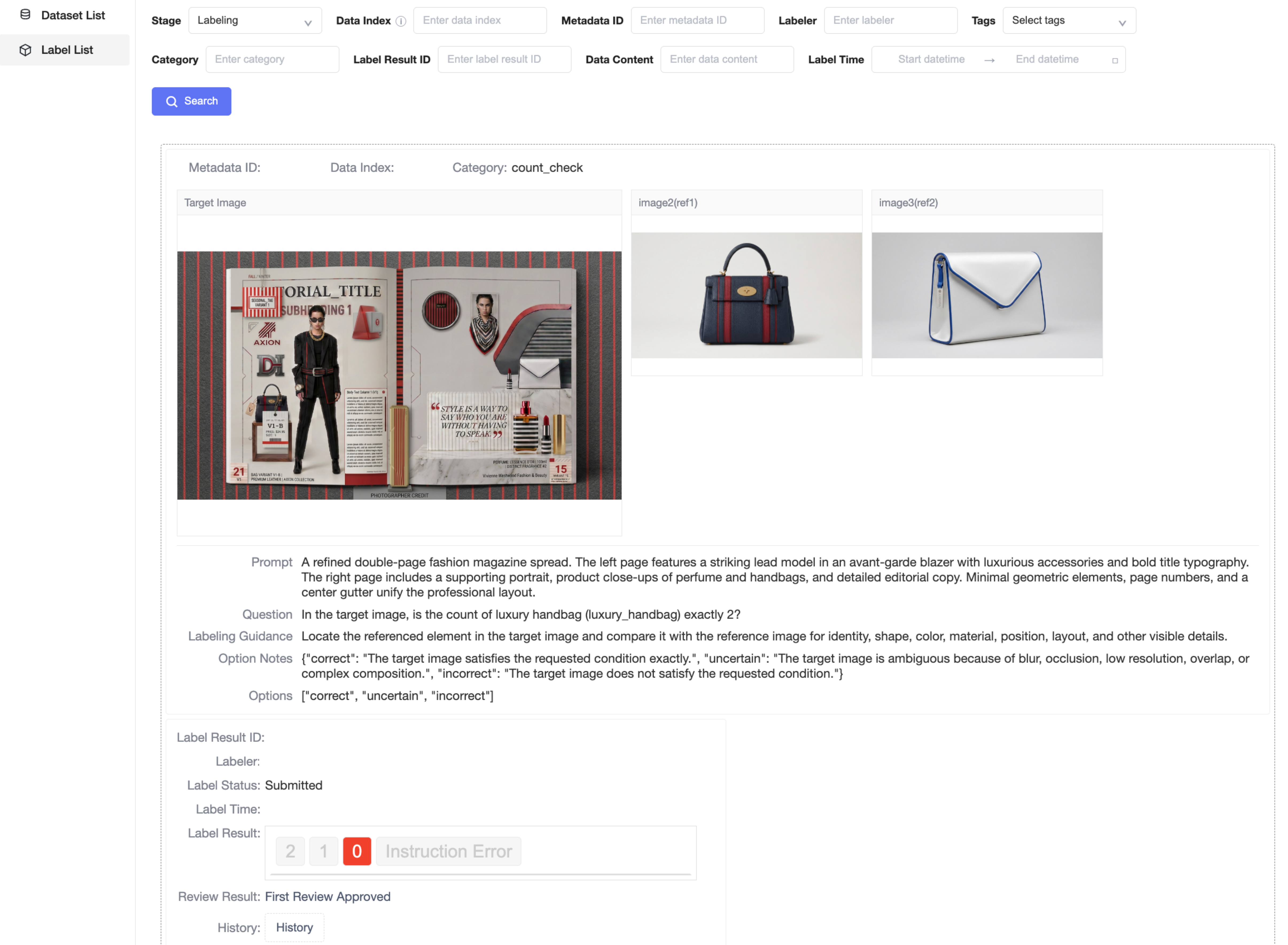}
  \par\vspace{5mm}
  \includegraphics[width=.96\linewidth]{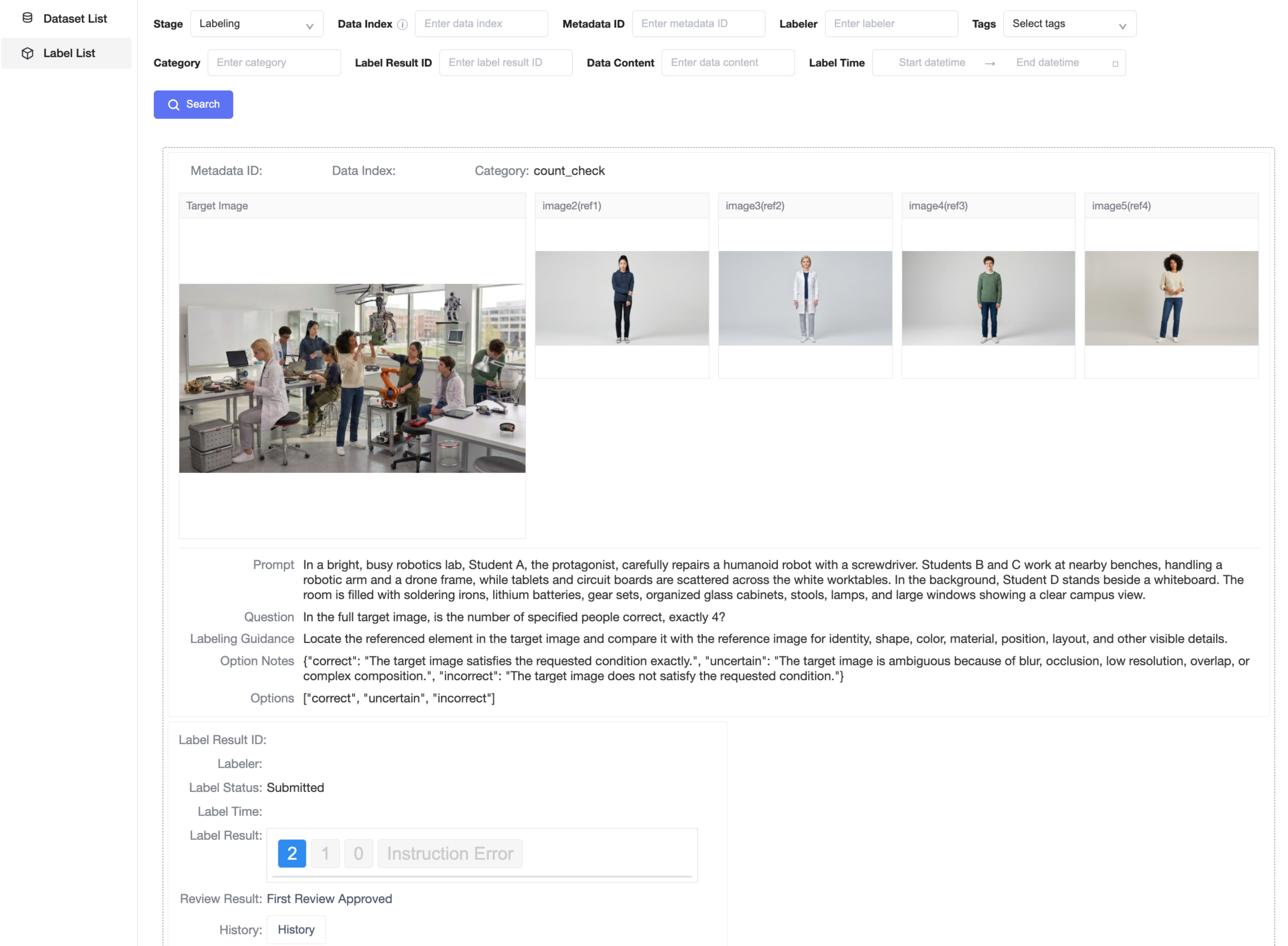}
  \caption{Human checklist annotation platform examples for count-oriented items. Top: a \texttt{count\_check} item asks whether the target image contains exactly two luxury handbags, with the relevant handbag references, prompt, item question, scoring guidance, submitted score, and first-review approval visible in the same audit record. Bottom: a \texttt{count\_check} item asks whether the full target image contains exactly four specified people, again tying the target output, references, item text, label submission and review result to a single checklist record.}
  \label{fig:human_annotation_platform_examples}
\end{figure}

\begin{figure}[H]
  \centering
  \includegraphics[width=.96\linewidth]{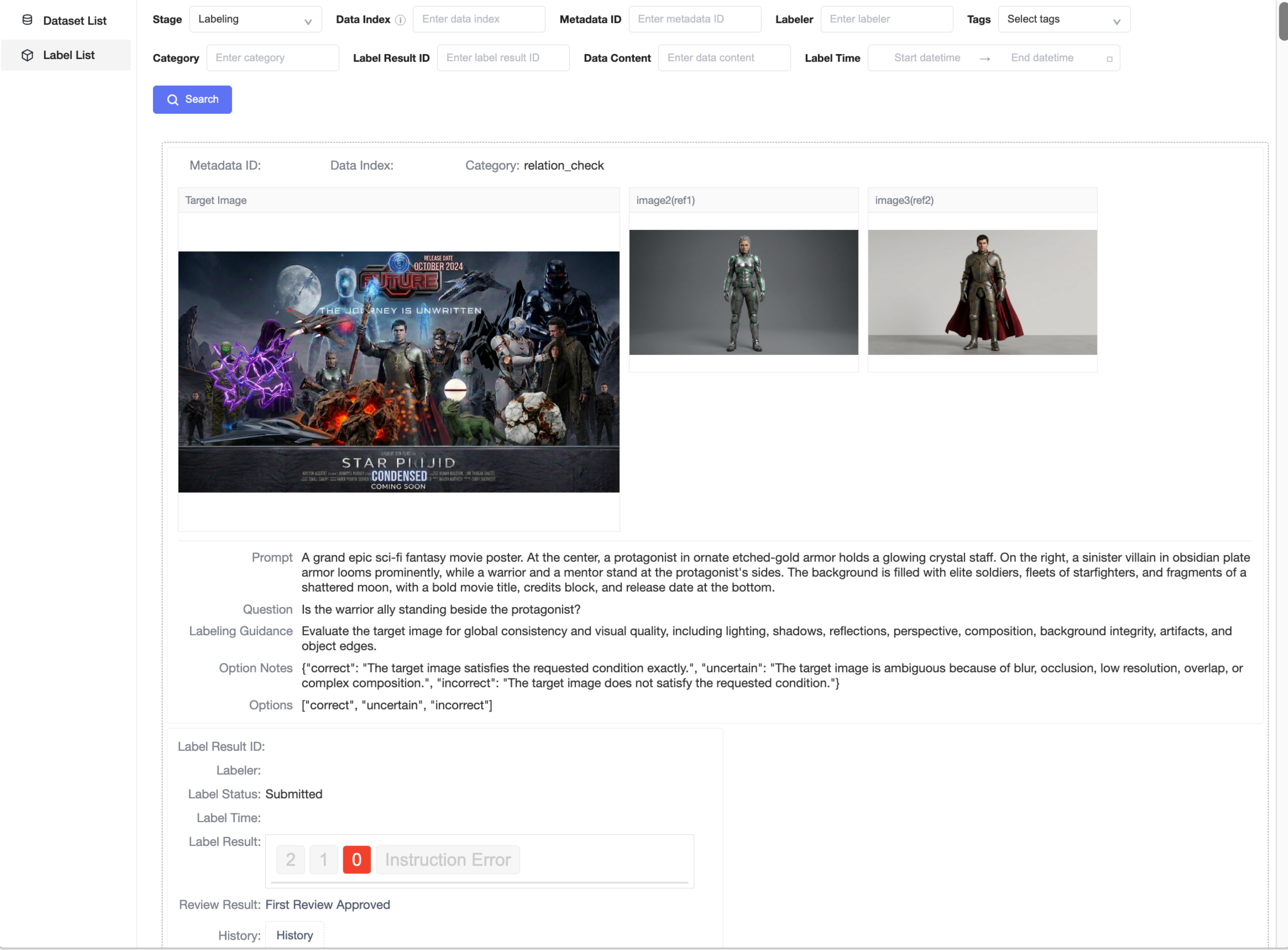}
  \caption{Human checklist annotation platform example for a relation item. The \texttt{relation\_check} screen asks whether the warrior ally stands beside the protagonist, requiring the reviewer to compare the target image with the relevant character references and apply the predefined ternary labels. The stored record includes metadata, category, guidance text, submitted score, and first-review approval, supporting strict post-hoc audit of each checklist decision.}
  \label{fig:human_annotation_platform_relation}
\end{figure}

\Cref{fig:human_annotation_platform_examples,fig:human_annotation_platform_relation} show representative checklist-item screens from the human annotation platform. Each item is evaluated as an atomic visual verification task rather than as a holistic image preference: the annotator sees the target image, the relevant reference images, the generation prompt, the item-specific question, labeling guidance, option definitions, and the allowed ternary labels. The platform records the case metadata, checklist category, label result identifier, annotator submission, timestamp, selected score, and review outcome. Final labels are accepted only after the review/adjudication process described above, yielding an auditable item-level trail for the 46{,}700 human-reviewed checklist labels used in the judge validation.

\clearpage

\clearpage
\begin{figure}[H]
  \centering
  \includegraphics[width=\linewidth]{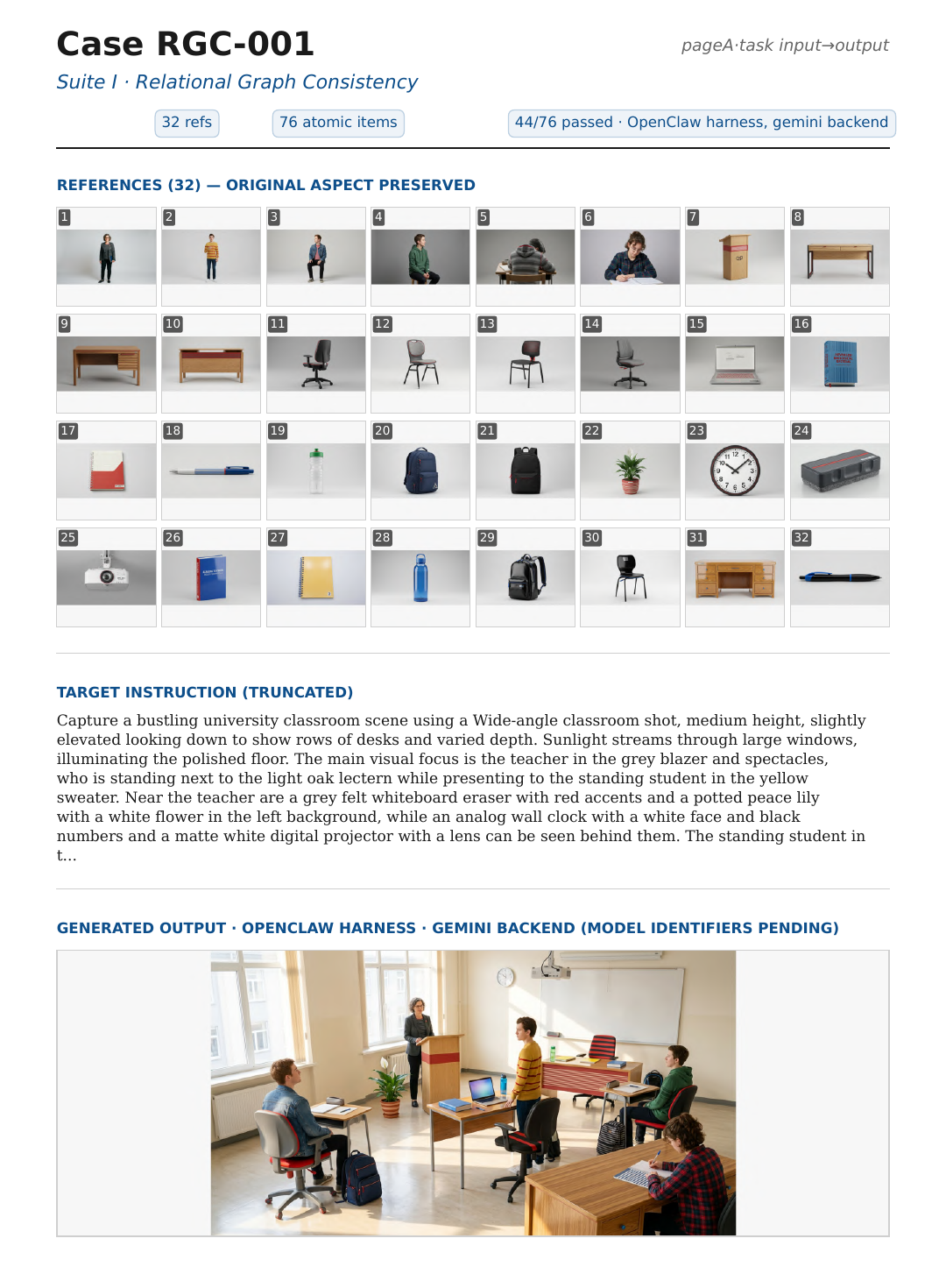}
  \caption{An example from the Relational Graph Consistency (RGC) subclass, with the input reference images, prompt, and a representative generated result.}
  \label{fig:rel_input}
\end{figure}

\begin{figure}[H]
  \centering
  \includegraphics[width=\linewidth]{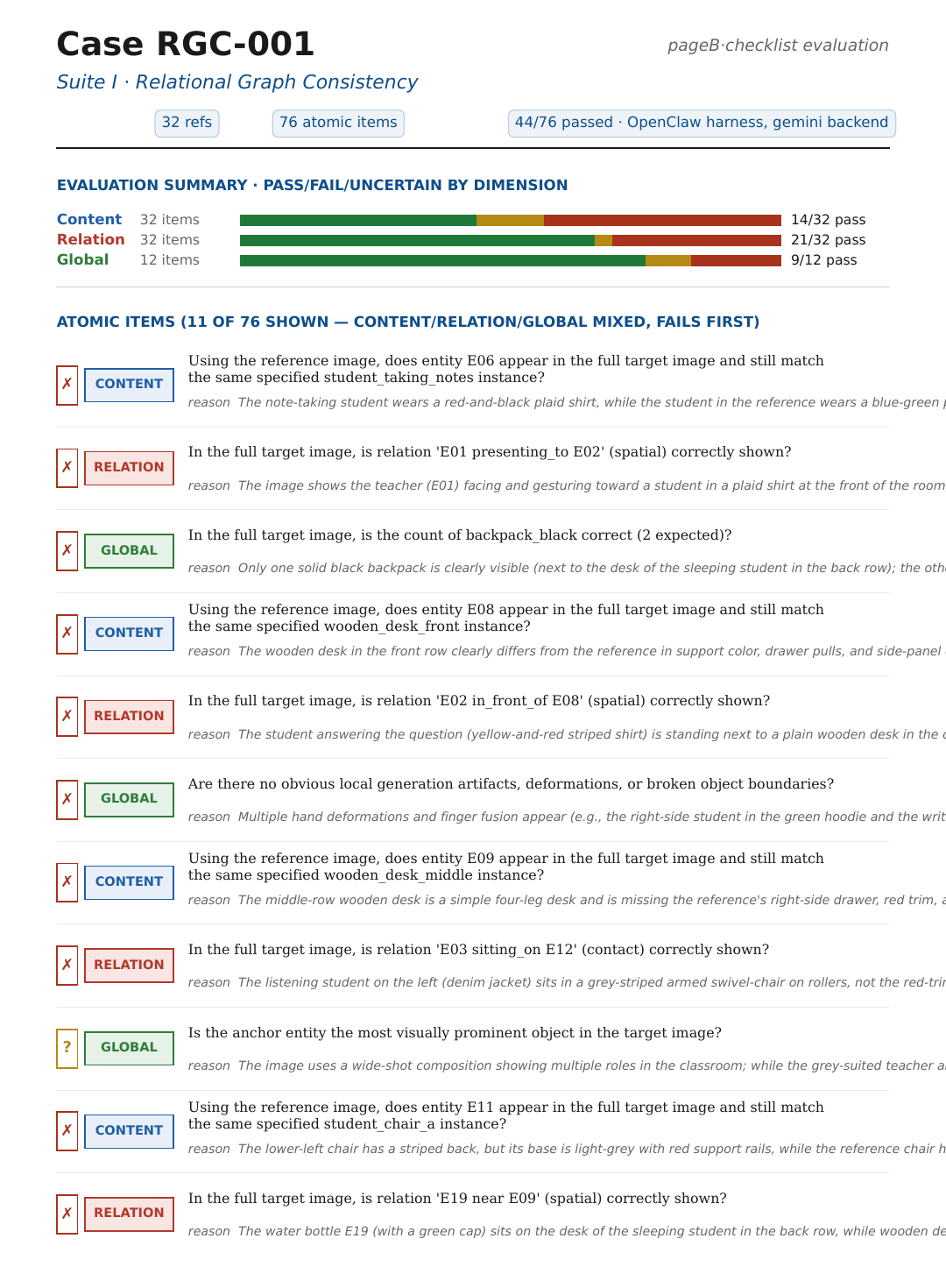}
  \caption{Partial evaluation checklist for an example from the Relational Graph Consistency (RGC) subclass.}
  \label{fig:rel_eval}
\end{figure}

\clearpage
\begin{figure}[H]
  \centering
  \includegraphics[width=\linewidth]{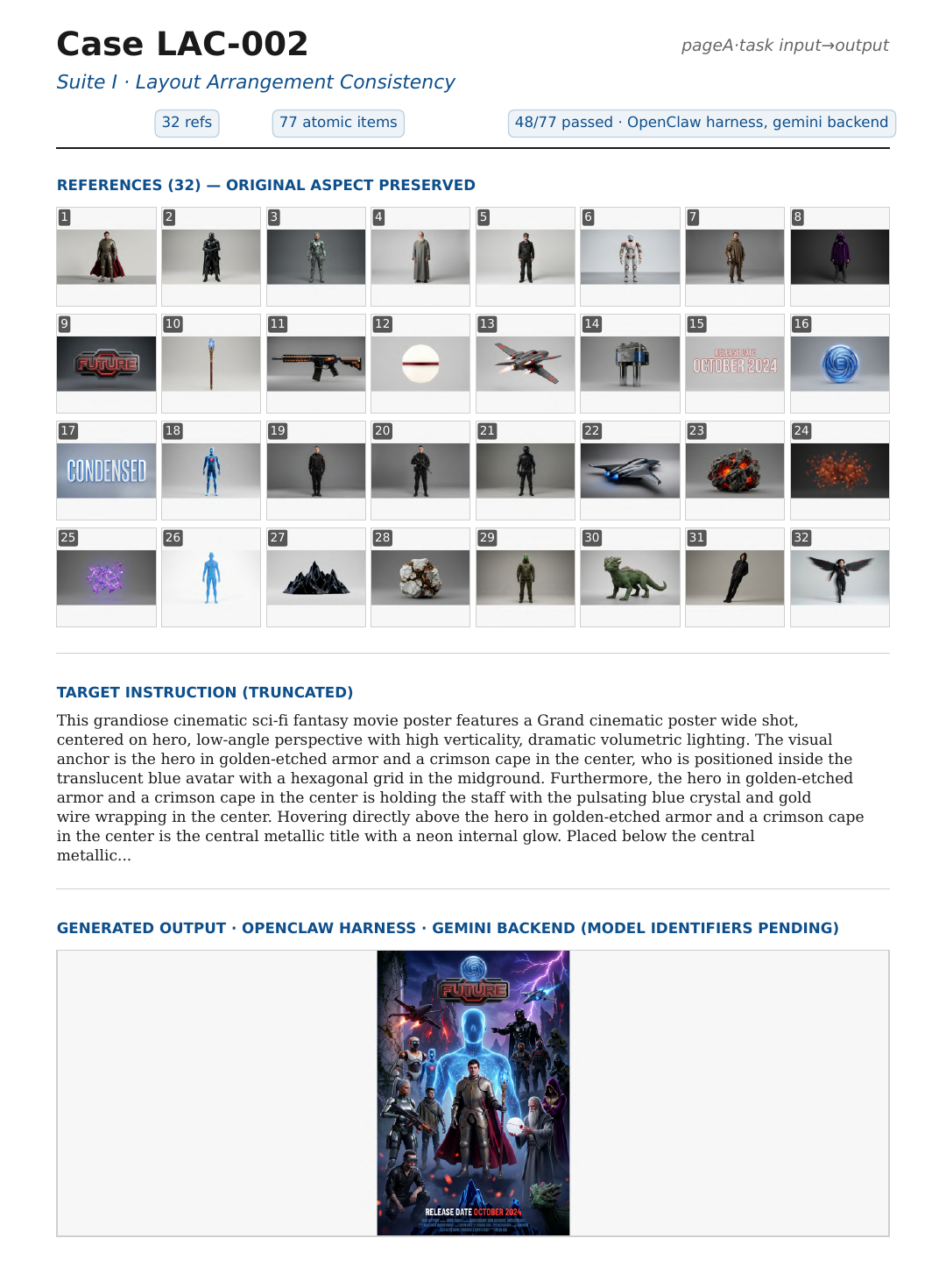}
  \caption{An example from the Layout Arrangement Consistency (LAC) subclass, with the input reference images, prompt, and a representative generated result.}
  \label{fig:lay_input}
\end{figure}

\begin{figure}[H]
  \centering
  \includegraphics[width=\linewidth]{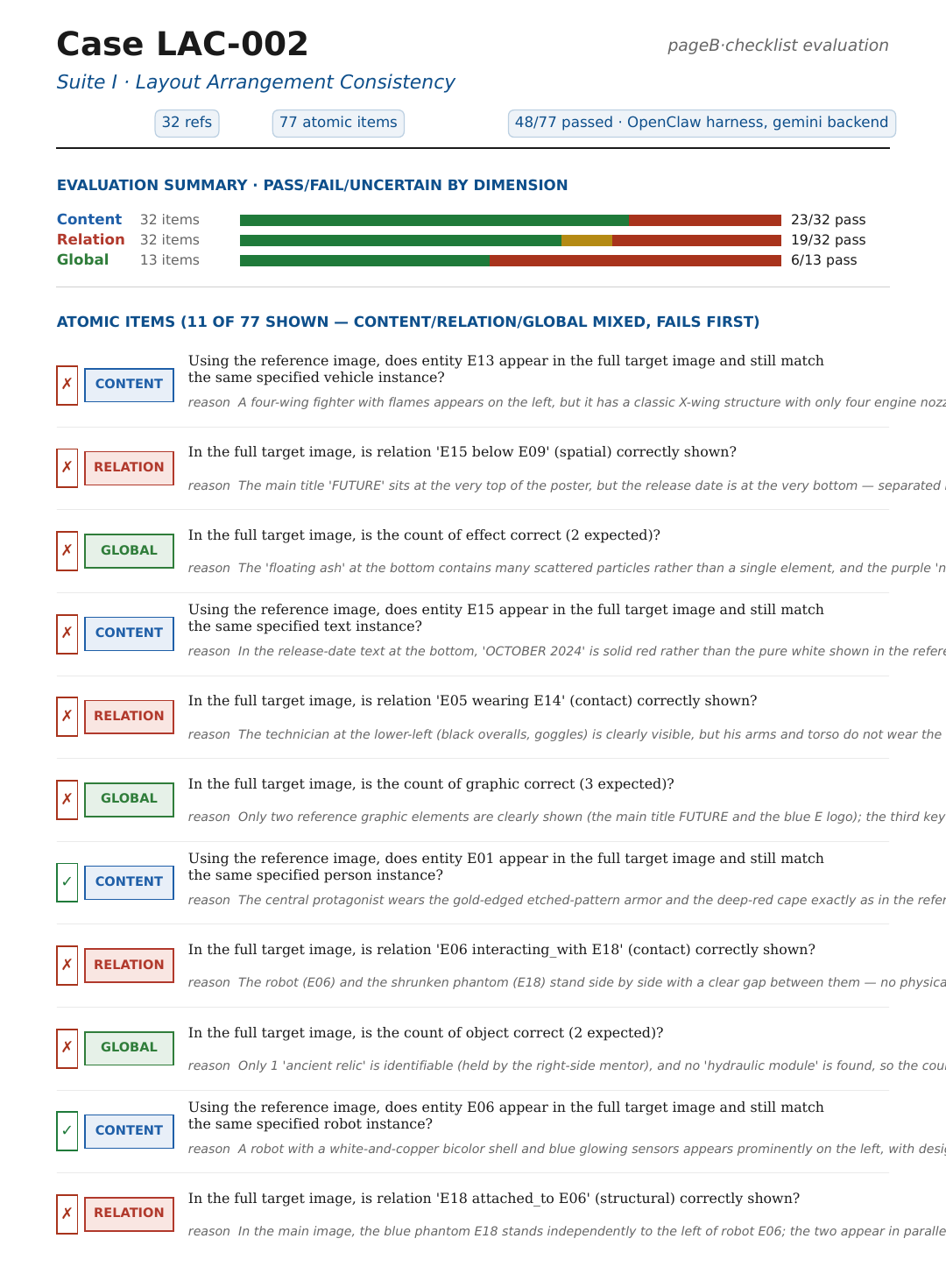}
  \caption{Partial evaluation checklist for an example from the Layout Arrangement Consistency (LAC) subclass.}
  \label{fig:lay_eval}
\end{figure}

\clearpage
\begin{figure}[H]
  \centering
  \includegraphics[width=\linewidth]{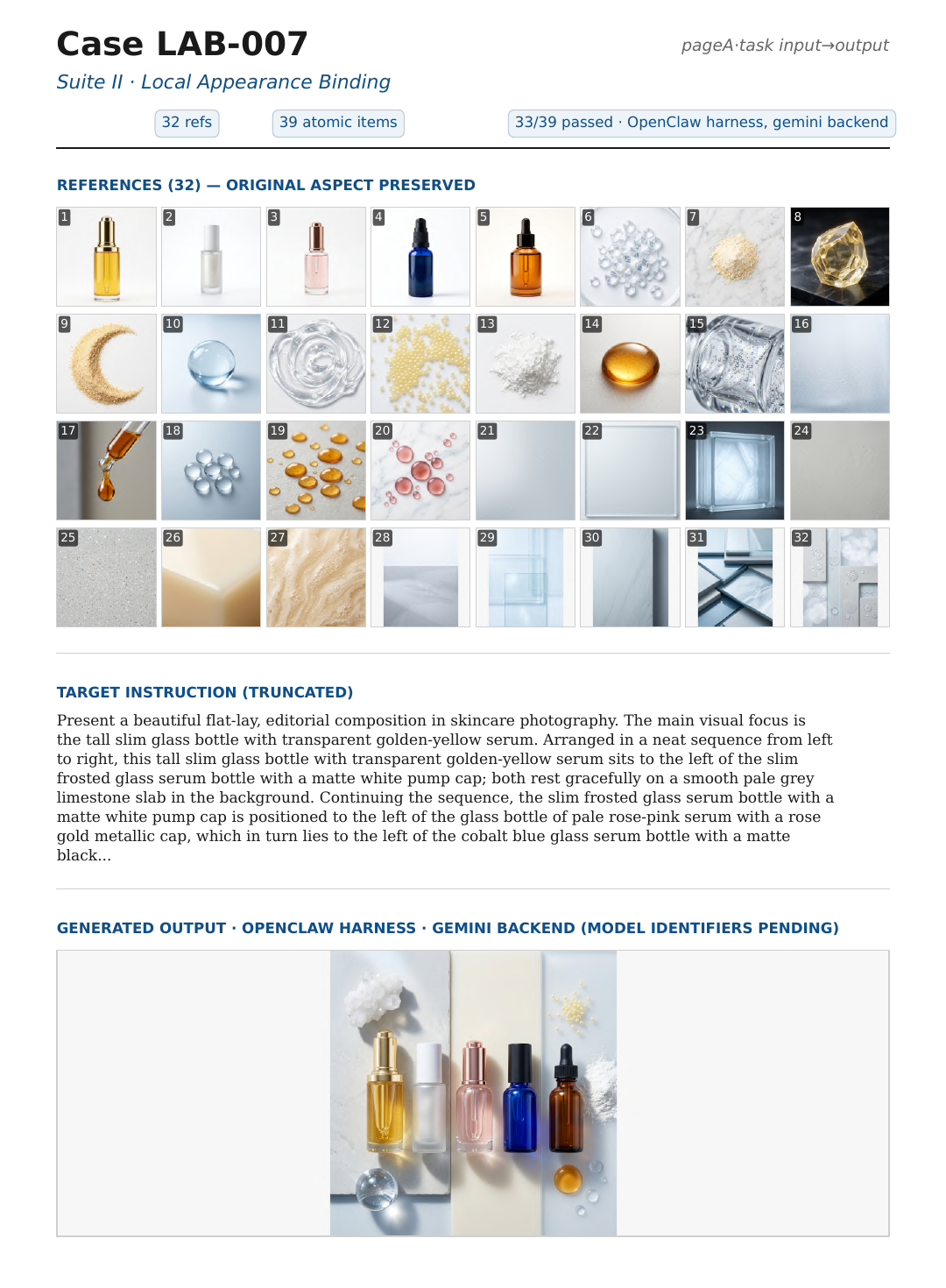}
  \caption{An example from the Local Appearance Binding (LAB) subclass, with the input reference images, prompt, and a representative generated result.}
  \label{fig:xab_input}
\end{figure}

\begin{figure}[H]
  \centering
  \includegraphics[width=\linewidth]{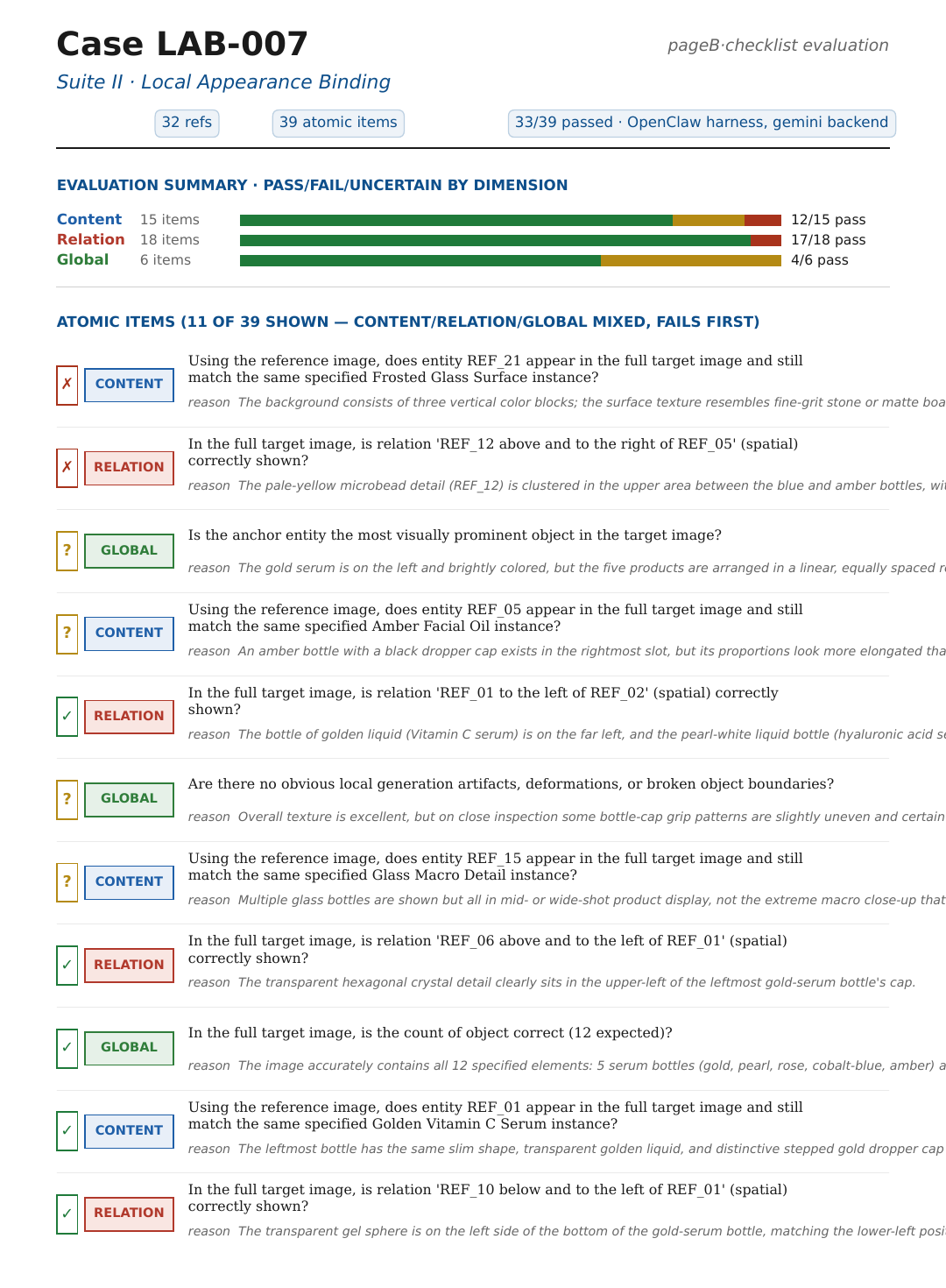}
  \caption{Partial evaluation checklist for an example from the Local Appearance Binding (LAB) subclass.}
  \label{fig:xab_eval}
\end{figure}

\clearpage
\begin{figure}[H]
  \centering
  \includegraphics[width=\linewidth]{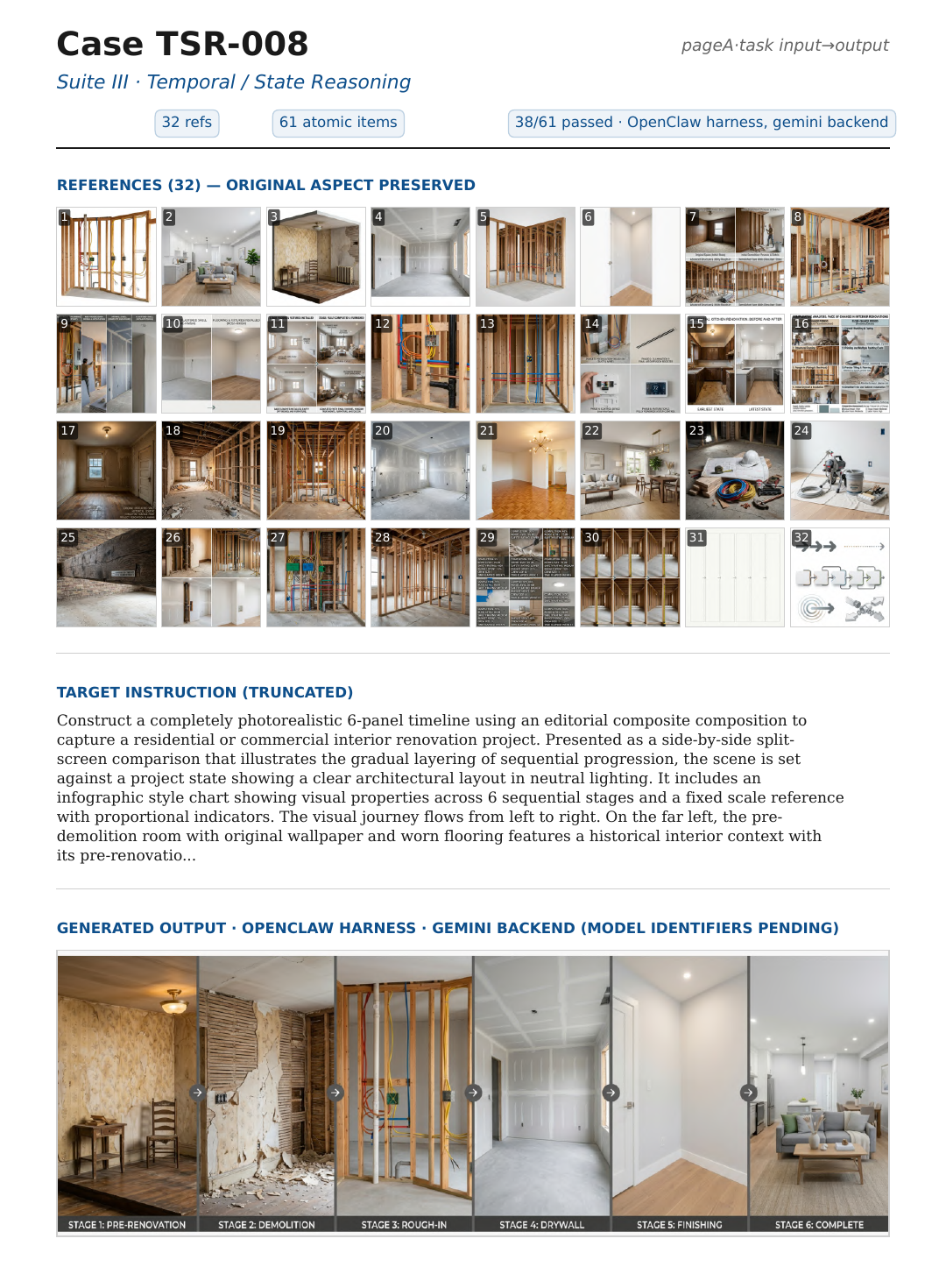}
  \caption{An example from the TSR task, with the input reference images, prompt and a representative generated result.}
  \label{fig:tsr_input}
\end{figure}

\begin{figure}[H]
  \centering
  \includegraphics[width=\linewidth]{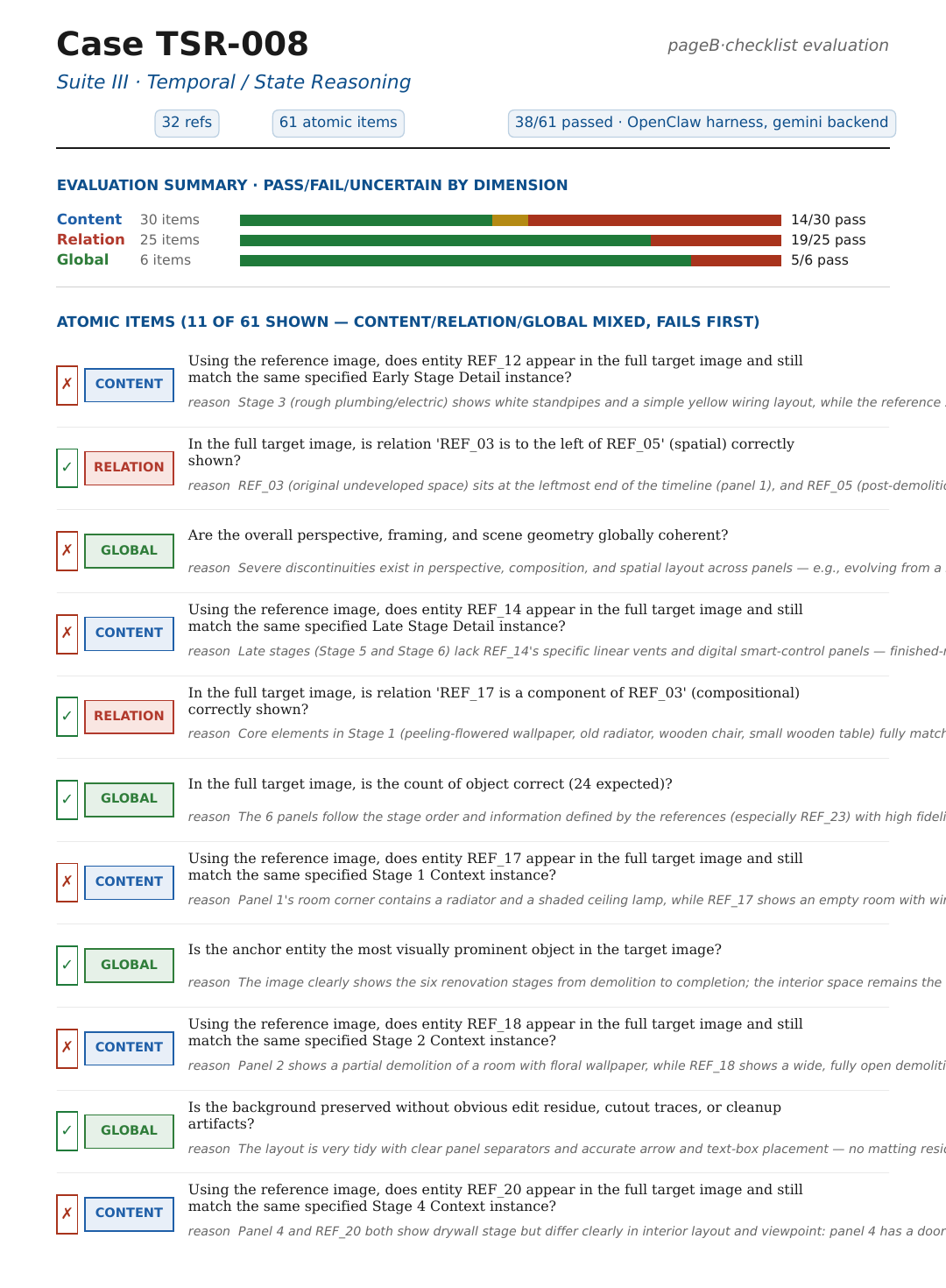}
  \caption{Partial evaluation checklist for an example from the TSR task.}
  \label{fig:tsr_eval}
\end{figure}

\clearpage
\begin{figure}[H]
  \centering
  \includegraphics[width=\linewidth]{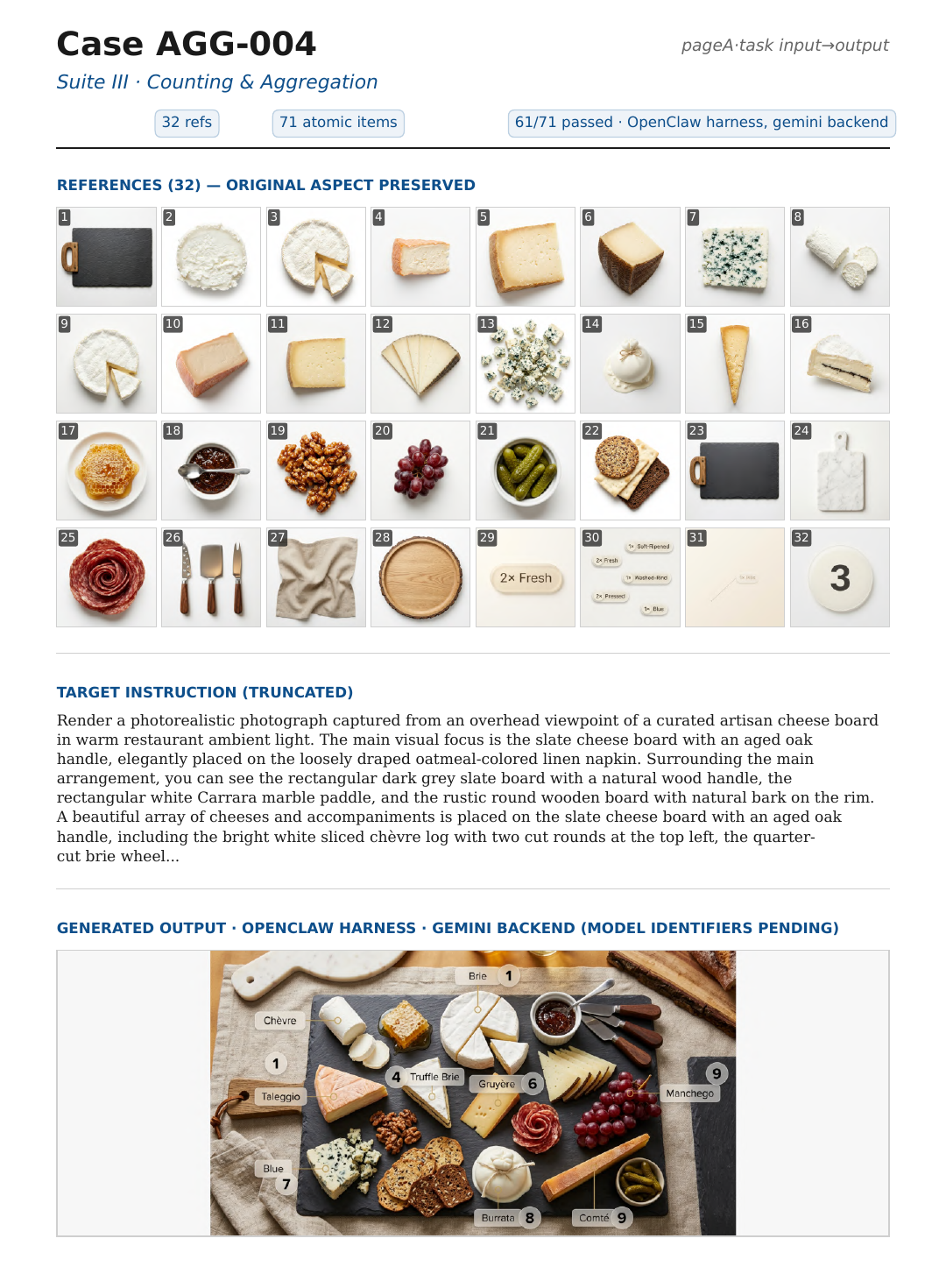}
  \caption{An example from the AGG task, with the input reference images, prompt and a representative generated result.}
  \label{fig:agg_input}
\end{figure}

\begin{figure}[H]
  \centering
  \includegraphics[width=\linewidth]{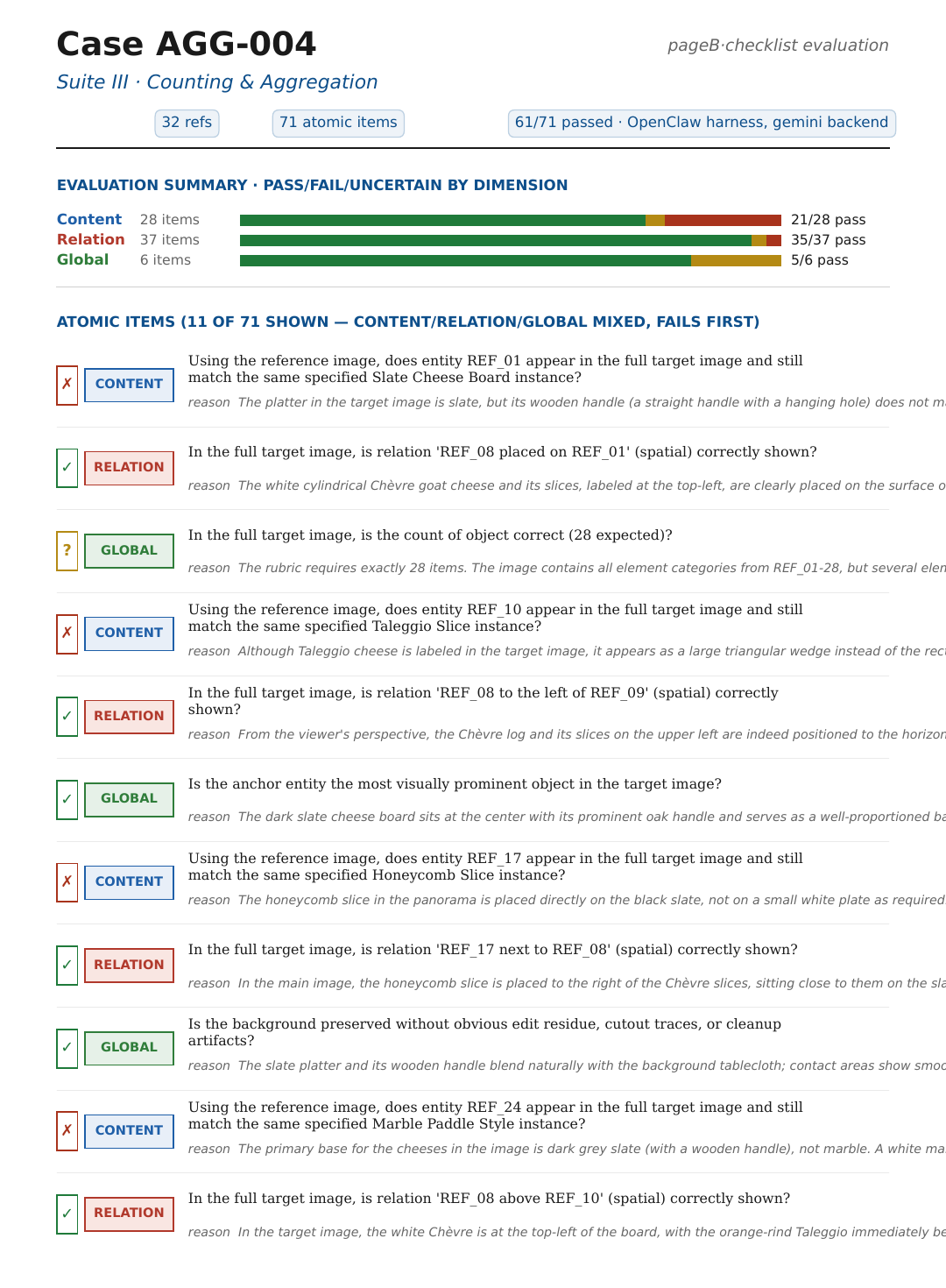}
  \caption{Partial evaluation checklist for an example from the AGG task.}
  \label{fig:agg_eval}
\end{figure}

\clearpage

\section{\ours{}-Eval Details}
\label{sec:suppl-eval}

\subsection{LLM and Agent Usage Declaration}
\scorer{} uses LLM or MLLM components as rubric critics, tool routers, and evidence-analysis modules. All automatic judging runs use Gemini 3 Flash as the shared judging backbone for both the agentic checklist judge and the single-pass VLM judge; in the agentic checklist judge, Gemini 3 Flash also drives tool routing and evidence analysis. \ours{}-Gen uses OpenCode, OpenClaw, and Hermes-Agent as harnesses. All three harnesses use Claude Sonnet 4.6 as the shared underlying agent model for planning, tool selection, and prompt revision. The harness versions are Hermes-Agent 0.10.0, OpenCode 1.4.9, and OpenClaw 2026.4.16. These harnesses call reference browsing, OCR/count/layout, image-generation, and file or prompt-management tools, while Nano Banana 2, GPT Image 1, or Seedream 4 produces image pixels according to the experimental setting.

\subsection{Human Agreement and Opposite-End Error Reduction}
The full-pool comparison uses all 46{,}700 reviewed checklist decisions from 900 generated outputs across 10 generation conditions. We compare VLM-as-judge with \scorer{} against the reviewed human labels. Each atomic check uses $2$ for satisfied/correct, $1$ for partially satisfied or uncertain, and $0$ for unsatisfied/incorrect; when absence is required, correct omission scores $2$. For a human label $y$ and evaluator label $\hat{y}$, exact ternary agreement requires $\hat{y}=y$, within-one agreement requires $|\hat{y}-y|\leq 1$, and opposite-end error occurs when $|\hat{y}-y|=2$.
\begin{table}[H]
  \centering
  \caption{Full-pool agreement with 46{,}700 reviewed human checklist labels from 900 generated outputs across 10 generation conditions. Higher is better for exact ternary and within-one agreement; lower is better for opposite-end error. Changes are \scorer{} minus VLM-as-judge, in percentage points (pp).}
  \label{tab:checklist_executor_control}
  \small
  \setlength{\tabcolsep}{4pt}
  \resizebox{\linewidth}{!}{  \begin{tabular}{lccc}
    \toprule
    Metric & VLM-as-judge & \scorer{} & Change \\
    \midrule
    Exact ternary agreement & 50.99\% & \textbf{66.81\%} & \textbf{+15.82 pp} \\
    Within-one agreement & 55.91\% & \textbf{75.19\%} & \textbf{+19.28 pp} \\
    Opposite-end $0 \leftrightarrow 2$ error & 44.09\% & \textbf{24.81\%} & \textbf{$-19.28$ pp} \\
    \bottomrule
  \end{tabular}}
\end{table}

\Cref{tab:checklist_executor_control} shows that exact ternary agreement increases by 15.82 percentage points, from 50.99\% to 66.81\%, while within-one agreement increases by 19.28 points, from 55.91\% to 75.19\%. Opposite-end $0 \leftrightarrow 2$ errors decrease from 44.09\% to 24.81\%, reducing complete satisfaction--failure reversals by 19.28 points. Within-one agreement and opposite-end error are complementary for ternary labels, so these two changes describe the same reduction in severe errors. These item-level results complement the output-level score correlations in \cref{tab:agreement_main} and support finer-grained diagnosis of reference-grounded failures.

\subsection{Tool Configuration and Aggregation}
The tool inventory covers OCR, object counting or detection, layout checking, and identity or feature matching. Tool calls produce observations rather than final labels. Item judgments use item-specific labels in $\{0,1,2\}$; for negative Content Fidelity items, absence is pass = 2, suspicious or ambiguous leakage is partial = 1, and full erroneous appearance is fail = 0. Checklist-average diagnostics multiply the mean label by 5, yielding a 0--10 score (rescaled to 0--100 for the subclass and Overall Avg. columns of the main generation tables) used for judge agreement and the main checklist-average columns; in the main generation tables, difficulty weighting is used only for the P@70\% column. No polarity inversion is applied after labeling.

\subsection{Difficulty-Weighted Generation Scoring}
\label{sec:suppl-generation-scoring}
The P@70\% column of the main generation tables uses a difficulty-weighted checklist score computed at the case--subclass task level; the subclass and Overall Avg. columns instead report the unweighted checklist-average score (the per-item mean rescaled to 0--100). Each checklist judgment is first mapped to a ternary item score $s_j\in\{0,1,2\}$: pass/match/correct is 2, partial satisfaction or uncertainty is 1, and fail/mismatch/incorrect/invalid is 0. Let $r_j=s_j/2$ be the normalized item score. We calibrate difficulty once using the reviewed human labels for the fixed reference pool of ten generation conditions. For each checklist item key $(\mathrm{case},\mathrm{subclass},\mathrm{item})$, $q_j$ is the mean normalized human label over this reference pool, defining the difficulty weight
\[
    w_j =
    \begin{cases}
      3, & q_j < 0.65,\\
      2, & 0.65 \le q_j < 0.90,\\
      1, & q_j \ge 0.90.
    \end{cases}
\]
The resulting weight map is saved as a versioned benchmark configuration and used unchanged in the main comparisons and for subsequent benchmark systems. The task score for strategy $a$ on case--subclass task $t$ is
\[
    \mathrm{TaskScore}(a,t)
    = 100\cdot\frac{\sum_{j\in t}w_js_{a,j}}{\sum_{j\in t}2w_j}
    = 100\cdot\frac{\sum_{j\in t}w_jr_{a,j}}{\sum_{j\in t}w_j}.
\]
P@70\% is the percentage of the 90 case--subclass tasks whose weighted score is at least 70; each benchmark case contributes one task in its primary subclass. The subclass and Overall Avg. columns of the main tables instead report the unweighted checklist-average score over the same tasks. For judge--human agreement, we separately use output-level checklist-average scores
\[
    s_o = 5\cdot\frac{1}{n_o}\sum_{j=1}^{n_o}y_{o,j},
    \qquad
    \hat{s}_o = 5\cdot\frac{1}{n_o}\sum_{j=1}^{n_o}\hat{y}_{o,j},
\]
to correlate reviewed labels $y$ with judge predictions $\hat{y}$ over the 900-output human-audited pool.

\clearpage

\section{Additional Generation Experiments and Diagnostics}
\label{sec:suppl-exp}

\begin{figure}[H]
  \centering
  \includegraphics[width=\linewidth]{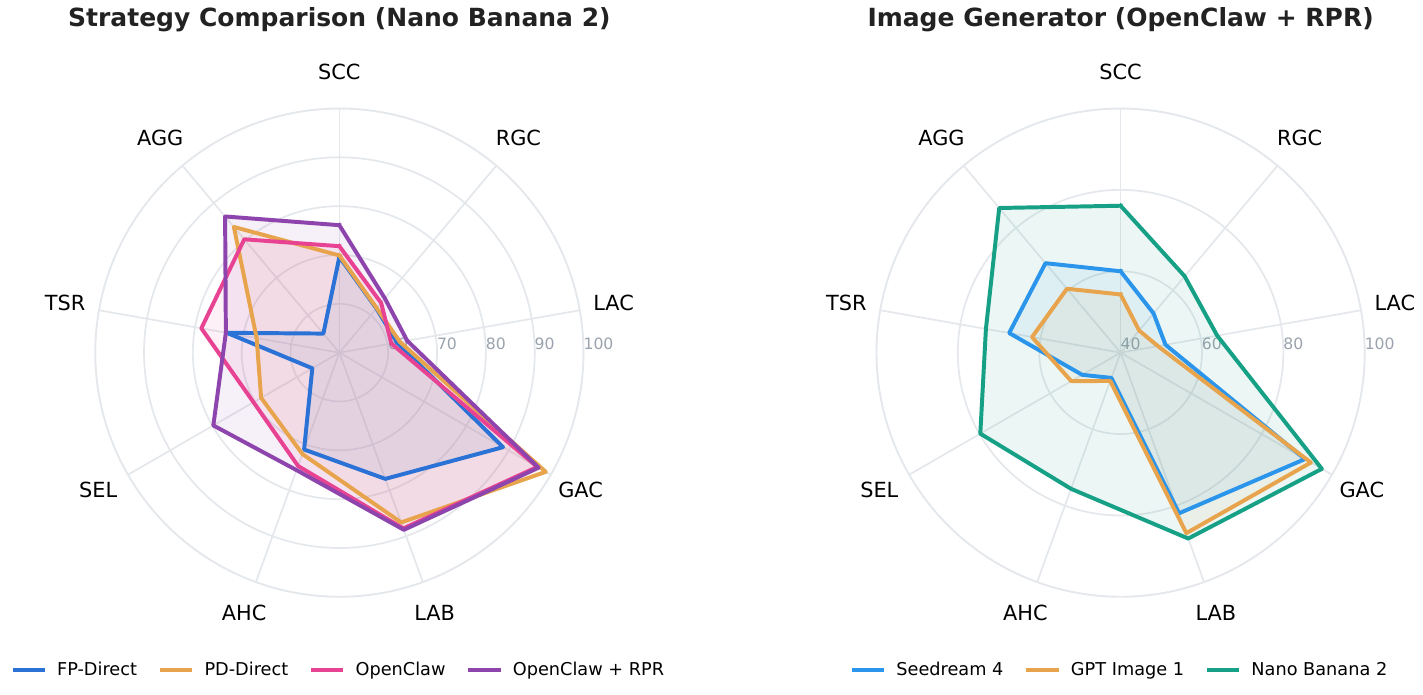}
  \caption{Radar profiles for the two primary result comparisons. Left: strategy comparison with Nano Banana 2 fixed as the image generator. Right: image-generator comparison with OpenClaw + RPR fixed as the harness. Axes are \ours{} taxonomy subclasses; values are the checklist-average subclass scores (0--100) reported in the main tables.}
  \label{fig:results_radar}
\end{figure}

\Cref{fig:results_radar} visualizes the two main result comparisons from \cref{tab:strategy_comparison,tab:generator_comparison_openclaw}. The radar profiles are intended as compact diagnostics of subclass-level failure patterns, not as additional statistical tests.

\subsection{Direct Baseline Implementations}
FP-Direct uses the full target prompt throughout generation and varies only the visible reference window. Because a case may contain more reference images than the backend can consume at once, the implementation uses a window of eight references and iterates over the ordered reference inventory, e.g., \texttt{ref1--ref8}, \texttt{ref9--ref16}, and so on until all non-background references have been exposed. Each round receives the same complete target prompt; there is no entity-level batching, graph-conditioned relation filtering, explicit state memory, or repair step.

PD-Direct is a deterministic staged-composition baseline rather than a free-form planning agent. It uses the same eight-reference backend budget as FP-Direct, but allocates that budget by graph structure rather than by inventory order. Each non-background entity node is represented by one reference image. Starting from the pre-generated background image, PD-Direct composes non-background entities in ordered batches of at most eight entity references, following the staging schedule that is optimal for the case's entity--relation graph. Let $b(v)\in\{1,\dots,K\}$ be the batch index of entity $v$, with $K=\lceil N/8\rceil$, at most eight entity references per batch, and the background fixed at batch $0$. Over the (unweighted, directed, possibly cyclic) entity--relation graph $G=(V,E)$, the schedule minimizes a lexicographic objective: first the number of relations $(u,v)$ whose anchor is composed later than its dependent, $\lvert\{(u,v)\in E: b(u)>b(v)\}\rvert$, a capacity-constrained feedback-arc count that we minimize exactly by integer programming, and, among schedules attaining that minimum, the total relational stretch $\sum_{(u,v)\in E}\lvert b(u)-b(v)\rvert$. We solve it exactly with an integer program (Pass~1 minimizes the feedback count $F^\star$; Pass~2 minimizes the stretch subject to the feedback count equal to $F^\star$); each case has a small number of composed entities ($N\le 32$), so the integer program solves in seconds. At step zero the backend receives the background image plus the first batch's reference images and produces \texttt{intermediate\_00.png}; at each later step it receives the previous intermediate image plus the next batch's references, and the final intermediate is saved as \texttt{target/target\_image.png}. For a case with $N=32$ composed entities this yields $K=\lceil N/8\rceil=4$ batches and four intermediate images. The prompt at each step is constructed deterministically from fixed staged-composition instructions, the scene prompt, the entity types in the current batch, and all graph edges incident to that batch (there is no per-step cap on edges, so every relation is presented at the step of its later endpoint), followed by constraints to preserve existing content and maintain consistent lighting, scale, and perspective. Reference images and the background are generated upstream from the entity catalog, scene specification, and entity--relation graph; PD-Direct only stages their use during target construction.

\subsection{Agent-Orchestrator Component Ablation}
\begin{figure}[H]
  \centering
  \includegraphics[width=\linewidth]{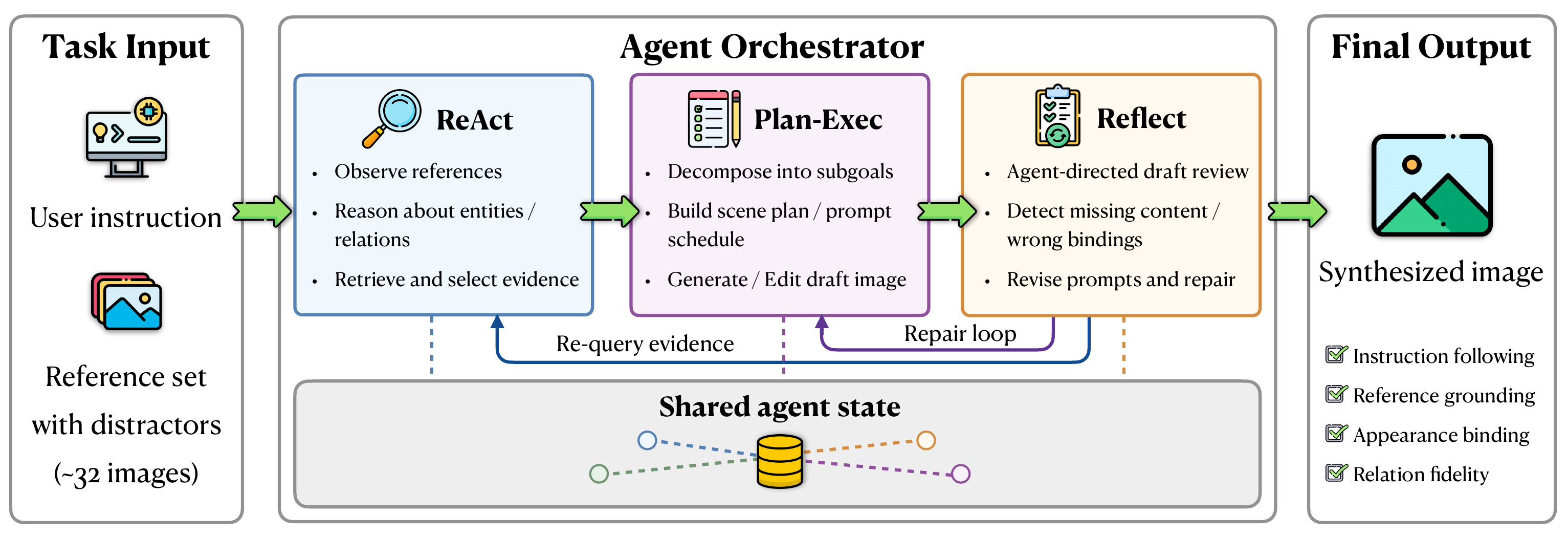}
  \caption{Agent-orchestrator workflow used in \ours{}-Gen. ReAct retrieves and reasons over reference evidence, Plan--Exec decomposes synthesis into executable subgoals, and Reflect inspects drafts to guide revision.}
  \label{fig:agent_orchestrator_appendix}
\end{figure}

\Cref{fig:agent_orchestrator_appendix} summarizes the ReAct, Plan--Exec, and Reflect stages used by the full RPR workflow. \Cref{tab:openclaw_component_ablation} isolates these stages with OpenClaw and Nano Banana 2 fixed, comparing the minimal OpenClaw harness, each single-stage variant, and the full OpenClaw + RPR stack. For this ablation, item difficulty is computed from the mean normalized item score across the five configurations in this table, using the thresholds in Appendix~\ref{sec:suppl-generation-scoring}. The resulting weights are shared across all five rows. Subclass scores, Overall Avg., and P@70\% therefore use this within-table relative difficulty calibration rather than the main benchmark's frozen reference-pool weights.

\begin{table}[H]
  \centering
  \caption{OpenClaw agent-generation component ablation on the 90-case \ours{} benchmark with Nano Banana 2 fixed as the image generator. Subclass scores and Overall Avg. are difficulty-weighted task scores (0--100), with difficulty calibrated across the five configurations shown; P@70\% is the corresponding task pass rate. Boldface and underlining mark the best and second-best values in each column.}
  \label{tab:openclaw_component_ablation}
  \scriptsize
  \setlength{\tabcolsep}{2.0pt}
  \begin{tabularx}{\linewidth}{@{}l*{11}{Y}@{}}
    \toprule
    & \multicolumn{3}{c}{Scene Composition}
    & \multicolumn{3}{c}{Visual Appearance}
    & \multicolumn{3}{c}{Reference Reasoning}
    & \multicolumn{2}{c}{Overall $\uparrow$} \\
    \cmidrule(lr){2-4}
    \cmidrule(lr){5-7}
    \cmidrule(lr){8-10}
    \cmidrule(lr){11-12}
    Strategy
    & SCC $\uparrow$ & RGC $\uparrow$ & LAC $\uparrow$
    & GAC $\uparrow$ & LAB $\uparrow$ & AHC $\uparrow$
    & SEL $\uparrow$ & TSR $\uparrow$ & AGG $\uparrow$
    & Avg. $\uparrow$ & \makebox[0pt][c]{P@70\% $\uparrow$} \\
    \midrule
    OpenClaw & 58.6 & 51.6 & 47.8 & 95.4 & 83.3 & 65.6 & 59.6 & \textbf{69.4} & 72.5 & \underline{67.1} & 40.0 \\
    Plan--Exec only & 56.6 & 45.7 & 46.5 & \textbf{96.5} & \textbf{88.3} & 63.7 & 59.0 & 58.4 & 66.8 & 64.6 & 35.6 \\
    ReAct only & 54.5 & 48.1 & 41.5 & 92.0 & 71.8 & 52.1 & \underline{62.0} & 57.0 & 72.8 & 61.3 & 36.7 \\
    Reflect only & \underline{58.7} & 46.4 & \textbf{50.6} & \underline{96.4} & 79.8 & 52.0 & 59.7 & \underline{64.4} & \underline{75.7} & 64.9 & \underline{44.4} \\
    OpenClaw + RPR & \textbf{64.3} & \textbf{54.5} & \underline{50.3} & 95.9 & \underline{83.8} & \textbf{66.6} & \textbf{72.6} & 62.0 & \textbf{79.6} & \textbf{70.0} & \textbf{47.8} \\
    \bottomrule
  \end{tabularx}
\end{table}

The ablation shows that the three agentic stages are complementary rather than individually monotonic. Plan--Exec alone gives the strongest Visual Appearance scores on GAC and LAB, consistent with decomposition helping product, style, and material organization, but it weakens relation control and structured reasoning when evidence retrieval and repair are not coupled to the plan. ReAct alone improves selection relative to the minimal OpenClaw harness, but its overall score drops because evidence selection without explicit execution control does not reliably preserve spatial and appearance bindings through generation. Reflect alone raises LAC, AGG, and P@70\%, indicating that revision helps repair visible layout and aggregation errors, but it can also over-correct or drift when missing evidence is not re-grounded.

The full RPR stack gives the best overall score and pass rate because the stages close each other's failure modes: ReAct grounds the reference set, Plan--Exec turns the grounded state into an executable synthesis schedule, and Reflect supports revision based on the generated result. The remaining TSR drop relative to minimal OpenClaw suggests that temporal and state-reasoning cases are sensitive to over-revision: once a physically plausible final state is found, additional repair can shift pose or event timing away from the intended state. Thus, RPR is best interpreted as an evidence-preserving orchestration pattern, not as a uniformly beneficial amount of extra reasoning.

\subsection{Case-Level Uncertainty and Paired Inference on the Checklist-Average Score}
\begin{table}[H]
  \centering
  \caption{Checklist-average generation summary with case-level uncertainty and paired-inference diagnostics. Scores use the same 0--100 scale as the main generation tables; mean differences are computed from the reported overall scores and expressed in percentage points (pp).}
  \label{tab:uncertainty_paired}
  \normalsize
  \setlength{\tabcolsep}{4pt}
  \renewcommand{\arraystretch}{0.95}
  \begin{tabular*}{\linewidth}{@{\extracolsep{\fill}}lccc@{}}
    \toprule
    Comparison & Score / diff. & Bootstrap 95\% CI & Paired $p$ \\
    \midrule
    \multicolumn{4}{c}{\itshape Minimal harness vs.\ RPR, with Nano Banana 2 fixed as backend} \\
    \midrule
    Hermes-Agent: RPR $-$ minimal & +1.7 pp & [+0.1, +3.7] pp & 0.0239 \\
    OpenCode: RPR $-$ minimal & +0.5 pp & [-1.3, +2.1] pp & 0.1823 \\
    OpenClaw: RPR $-$ minimal & +2.3 pp & [+0.6, +4.2] pp & 0.0032 \\
    \midrule
    \multicolumn{4}{c}{\itshape Image backend comparison, with OpenClaw + RPR fixed as the harness} \\
    \midrule
    Nano Banana 2 Overall & 78.4 & [76.7, 80.1] & -- \\
    Seedream 4 Overall & 63.6 & [61.5, 65.6] & -- \\
    GPT Image 1 Overall & 61.8 & [59.5, 64.0] & -- \\
    Nano Banana 2 $-$ Seedream 4 paired diff. & +14.8 pp & [+12.9, +16.8] pp & $\approx 10^{-5}$ \\
    Nano Banana 2 $-$ GPT Image 1 paired diff. & +16.6 pp & [+14.5, +18.8] pp & $\approx 10^{-5}$ \\
    \midrule
    \multicolumn{4}{c}{\itshape Suite scores for OpenClaw + RPR + Nano Banana 2} \\
    \midrule
    Scene Composition & 68.2 & [65.7, 70.7] & -- \\
    Visual Appearance & 87.1 & [83.9, 90.1] & -- \\
    Reference Reasoning & 79.9 & [76.6, 82.9] & -- \\
    \bottomrule
  \end{tabular*}
\end{table}

\clearpage

\section{Responsible Release and Safeguards}
\label{sec:suppl-responsible-release}

\subsection{Limitations}
The $3 \times 3$ taxonomy covers a meaningful subset of long-horizon multi-reference synthesis, but does not cover video generation, 3D scene composition, or unrestricted multimodal planning. The benchmark is intentionally high-density, which improves diagnostic resolution but makes expert annotation expensive. \scorer{} depends on backbone multimodal models and verification tools, so model identifiers and tool versions must accompany final results.

\subsection{Broader Impact Details}
The benchmark is designed for measurement rather than deployment, but it evaluates systems capable of generating realistic composite images from many references. Positive uses include model auditing and failure diagnosis. Risks include deceptive media, impersonation, privacy leakage, biased reference coverage, and copyright-sensitive reuse. Mitigations include filtering unsafe content, documenting provenance, restricting problematic assets, and providing a project issue tracker for reporting release problems at \url{https://github.com/TIGER-AI-Lab/LongBanana/issues}.

\newpage
\section*{NeurIPS Paper Checklist}

\begin{enumerate}

\item {\bf Claims}
    \item[] Question: Do the main claims made in the abstract and introduction accurately reflect the paper's contributions and scope?
    \item[] Answer: \answerYes{}
    \item[] Justification: The abstract and \cref{sec:intro} state the benchmark, evaluator, and empirical claims, while \cref{sec:benchmark,sec:evaluator,sec:exp,sec:conclusion} define the scope and required evidence. Run-specific factual details are provided in the anonymized OpenReview artifacts and associated run records.

\item {\bf Limitations}
    \item[] Question: Does the paper discuss the limitations of the work performed by the authors?
    \item[] Answer: \answerYes{}
    \item[] Justification: Appendix~\ref{sec:suppl-responsible-release} discusses benchmark scope, annotation expense, and evaluator dependence.

\item {\bf Theory assumptions and proofs}
    \item[] Question: For each theoretical result, does the paper provide the full set of assumptions and a complete (and correct) proof?
    \item[] Answer: \answerNA{}
    \item[] Justification: The paper does not present theoretical theorems or formal proof claims.

    \item {\bf Experimental result reproducibility}
    \item[] Question: Does the paper fully disclose all the information needed to reproduce the main experimental results of the paper to the extent that it affects the main claims and/or conclusions of the paper (regardless of whether the code and data are provided or not)?
    \item[] Answer: \answerNo{}
    \item[] Justification: The paper describes benchmark construction, scoring, model and harness identifiers, and baseline implementations in \cref{sec:benchmark,sec:evaluator,sec:exp,sec:suppl-bench,sec:suppl-eval,sec:suppl-exp}, but does not include executable reproduction commands or environment files.

\item {\bf Open access to data and code}
    \item[] Question: Does the paper provide open access to the data and code, with sufficient instructions to faithfully reproduce the main experimental results, as described in supplemental material?
    \item[] Answer: \answerNo{}
    \item[] Justification: This preprint does not provide a public data/code access link with executable reproduction instructions.

\item {\bf Experimental setting/details}
    \item[] Question: Does the paper specify all the training and test details (e.g., data splits, hyperparameters, how they were chosen, type of optimizer) necessary to understand the results?
    \item[] Answer: \answerYes{}
    \item[] Justification: The experimental design, 90-case benchmark, evaluator configuration, metrics, and ablations are described in \cref{sec:benchmark,sec:evaluator,sec:exp}. Appendix~\ref{sec:suppl-eval} identifies the agent and judging models and the harness versions.

\item {\bf Experiment statistical significance}
    \item[] Question: Does the paper report error bars suitably and correctly defined or other appropriate information about the statistical significance of the experiments?
    \item[] Answer: \answerYes{}
    \item[] Justification: \Cref{sec:exp} reports checklist-average generation scores and the secondary difficulty-weighted P@70\% diagnostic. Appendix~\ref{sec:suppl-exp} reports case-level bootstrap confidence intervals and paired tests for the checklist-average scores; judge--human agreement is reported as correlation diagnostics.

\item {\bf Experiments compute resources}
    \item[] Question: For each experiment, does the paper provide sufficient information on the computer resources (type of compute workers, memory, time of execution) needed to reproduce the experiments?
    \item[] Answer: \answerNo{}
    \item[] Justification: The paper identifies the generation and judging models but does not report per-run compute resources, runtime, or API costs.
    
\item {\bf Code of ethics}
    \item[] Question: Does the research conducted in the paper conform, in every respect, with the NeurIPS Code of Ethics \url{https://neurips.cc/public/EthicsGuidelines}?
    \item[] Answer: \answerYes{}
    \item[] Justification: Asset provenance and privacy screening are discussed in Appendix~\ref{sec:suppl-bench}; broader impacts and responsible-release safeguards are discussed in Appendix~\ref{sec:suppl-responsible-release}.

\item {\bf Broader impacts}
    \item[] Question: Does the paper discuss both potential positive societal impacts and negative societal impacts of the work performed?
    \item[] Answer: \answerYes{}
    \item[] Justification: Appendix~\ref{sec:suppl-responsible-release} discusses positive impacts for evaluation rigor and negative risks related to deceptive media, privacy, copyright, and biased coverage, together with mitigation measures.
    
\item {\bf Safeguards}
    \item[] Question: Does the paper describe safeguards that have been put in place for responsible release of data or models that have a high risk for misuse (e.g., pre-trained language models, image generators, or scraped datasets)?
    \item[] Answer: \answerYes{}
    \item[] Justification: Appendix~\ref{sec:suppl-bench} describes asset screening and redistribution restrictions; Appendix~\ref{sec:suppl-responsible-release} describes responsible-release safeguards and takedown/contact procedures.

\item {\bf Licenses for existing assets}
    \item[] Question: Are the creators or original owners of assets (e.g., code, data, models), used in the paper, properly credited and are the license and terms of use explicitly mentioned and properly respected?
    \item[] Answer: \answerYes{}
    \item[] Justification: Asset provenance and licensing are documented in \cref{sec:benchmark,sec:suppl-bench}, covering source images, models, tools, and external benchmarks. Source citations, license/TOS audit results, and redistribution decisions are included in the submitted artifact metadata.

\item {\bf New assets}
    \item[] Question: Are new assets introduced in the paper well documented and is the documentation provided alongside the assets?
    \item[] Answer: \answerYes{}
    \item[] Justification: The benchmark, manifests, annotation schema, scorer configuration, and documentation are described in \cref{sec:benchmark,sec:evaluator,sec:suppl-bench,sec:suppl-eval}. The dataset card, manifest, release license, and anonymized artifact access information are provided in the submitted OpenReview artifacts.

\item {\bf Crowdsourcing and research with human subjects}
    \item[] Question: For crowdsourcing experiments and research with human subjects, does the paper include the full text of instructions given to participants and screenshots, if applicable, as well as details about compensation (if any)? 
    \item[] Answer: \answerYes{}
    \item[] Justification: The annotation workflow, adjudication plan, compensation reporting, and instruction/screenshot requirements are described in \cref{sec:benchmark,sec:suppl-bench}. Instructions, UI screenshots, rater counts, qualification criteria, compensation rate, and annotation time are reported as annotation metadata.

\item {\bf Institutional review board (IRB) approvals or equivalent for research with human subjects}
    \item[] Question: Does the paper describe potential risks incurred by study participants, whether such risks were disclosed to the subjects, and whether Institutional Review Board (IRB) approvals (or an equivalent approval/review based on the requirements of your country or institution) were obtained?
    \item[] Answer: \answerYes{}
    \item[] Justification: Participant-risk and review-status reporting are specified in \cref{sec:suppl-bench,sec:suppl-responsible-release} while preserving submission anonymity. The IRB exemption or equivalent review status is reported in anonymized form where applicable.

\item {\bf Declaration of LLM usage}
    \item[] Question: Does the paper describe the usage of LLMs if it is an important, original, or non-standard component of the core methods in this research? Note that if the LLM is used only for writing, editing, or formatting purposes and does \emph{not} impact the core methodology, scientific rigor, or originality of the research, declaration is not required.
        \item[] Answer: \answerYes{}
    \item[] Justification: LLM and MLLM usage in generation harnesses, rubric critics, tool routing, and evidence analysis is described in \cref{sec:evaluator,sec:suppl-eval}. The appendix reports the shared underlying agent model for generation harnesses, harness versions, and the Gemini 3 Flash judging backbone used by both the agentic checklist judge and the single-pass VLM judge.

\end{enumerate}

\end{document}